\documentclass[10pt,twocolumn,letterpaper]{article}

\usepackage[pagenumbers]{cvpr} 

\usepackage{graphicx}
\usepackage{amsmath}
\usepackage{amssymb}
\usepackage{booktabs}
\usepackage{times}
\usepackage{epsfig}
\usepackage{caption}
\usepackage[ruled,vlined,linesnumbered]{algorithm2e}
\usepackage{enumitem}
\usepackage{multirow}
\usepackage[misc]{ifsym}

\usepackage{makecell}
\usepackage{cases}
\usepackage{algorithmic}
\usepackage{xcolor}
\usepackage{rotating}
\usepackage[pagebackref,breaklinks,colorlinks]{hyperref}

\usepackage[accsupp]{axessibility}

\usepackage[capitalize]{cleveref}
\crefname{section}{Sec.}{Secs.}
\Crefname{section}{Section}{Sections}
\Crefname{table}{Table}{Tables}
\crefname{table}{Tab.}{Tabs.}

\def\confName{CVPR}
\def\confYear{2022}

\newcommand{\nickname}{HCMoCo}

\begin{document}

\setlength{\abovedisplayskip}{3pt}
\setlength{\belowdisplayskip}{3pt}

\title{Versatile Multi-Modal Pre-Training for Human-Centric Perception}

\author{
    Fangzhou Hong\textsuperscript{1}, \quad
    Liang Pan\textsuperscript{1}, \quad
    Zhongang Cai\textsuperscript{1,2,3}, \quad
    Ziwei Liu\textsuperscript{1~\Letter} \\
    \textsuperscript{1}S-Lab, Nanyang Technological University \quad
    \textsuperscript{2}SenseTime Research \quad
    \textsuperscript{3}Shanghai AI Laboratory \\
    \texttt{\small \{fangzhou001, liang.pan, ziwei.liu\}@ntu.edu.sg} \quad \texttt{\small caizhongang@sensetime.com}
}

\twocolumn[{
    \renewcommand\twocolumn[1][]{#1}%
    \maketitle
    \vspace{-36pt}
    \begin{center}
        \centering
        \includegraphics[width=1.0\textwidth]{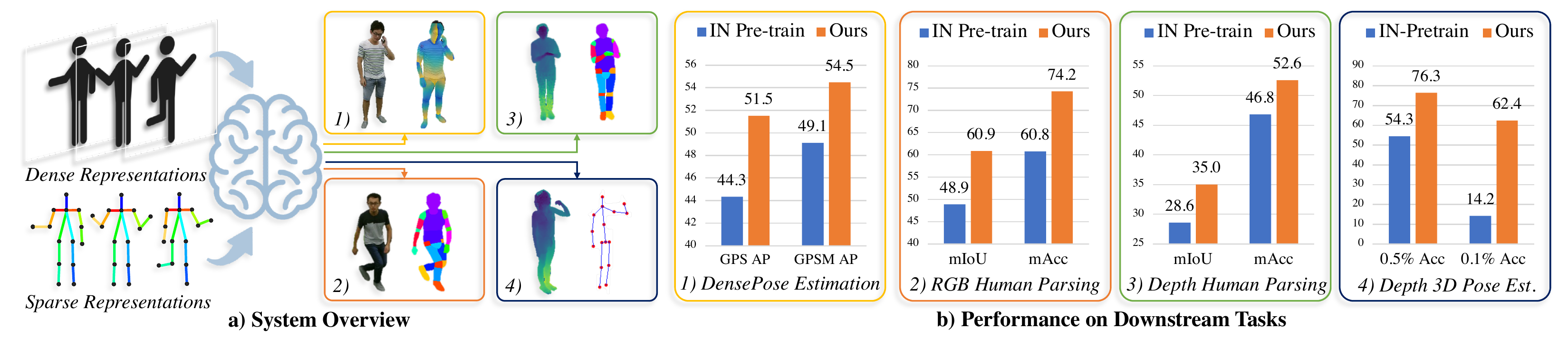}
        \vspace{-25pt}
        \captionof{figure}{
        \textbf{An Overview of \nickname{}.} a) We present \nickname{}, a versatile multi-modal pre-training framework that takes multi-modal observations of human body as input for human-centric perception.
        The pre-train models can be transferred to various human-centric downstream tasks with different modalities.
        b) Our \nickname{} shows superior performance on all four downstream tasks, especially for data-efficient settings ($10\%$ DensePose, $20\%$ RGB/depth human parsing, $0.5/0.1\%$ 3D pose estimation). `IN' stands for ImageNet.
        }
        \label{fig:teaser}
    \end{center}
    \vspace{-6pt}
}]

\maketitle

\setlength{\skip\footins}{12pt}
\newcommand\blfootnote[1]{%
\begingroup
\renewcommand\thefootnote{}\footnote{#1}%
\addtocounter{footnote}{-1}%
\endgroup
}
\blfootnote{\textsuperscript{\Letter}~Corresponding author}

\begin{abstract}    
    Human-centric perception plays a vital role in vision and graphics. But their data annotations are prohibitively expensive.
    Therefore, it is desirable to have a versatile pre-train model that serves as a foundation for data-efficient downstream tasks transfer.
    To this end, we propose the \textbf{H}uman-\textbf{C}entric Multi-\textbf{Mo}dal \textbf{Co}ntrastive Learning framework \textbf{\nickname{}} that leverages the multi-modal nature of human data (\eg RGB, depth, 2D keypoints) for effective representation learning.
    The objective comes with two main challenges: dense pre-train for multi-modality data, efficient usage of sparse human priors.
    To tackle the challenges, we design the novel Dense Intra-sample Contrastive Learning and Sparse Structure-aware Contrastive Learning targets by hierarchically learning a modal-invariant latent space featured with continuous and ordinal feature distribution and structure-aware semantic consistency.
    \nickname{} provides pre-train for different modalities by combining heterogeneous datasets, which allows efficient usage of existing task-specific human data.
    Extensive experiments on four downstream tasks of different modalities demonstrate the effectiveness of \nickname{}, especially under data-efficient settings (7.16\% and 12\% improvement on DensePose Estimation and Human Parsing). 
    Moreover, we demonstrate the versatility of \nickname{} by exploring cross-modality supervision and missing-modality inference, validating its strong ability in cross-modal association and reasoning. Codes are available at \url{https://github.com/hongfz16/HCMoCo}.
\end{abstract}

\vspace{-18pt}
\section{Introduction}
\label{sec:intro}

As a long-standing problem, human-centric perception has been studied for decades, ranging from sparse prediction tasks, such as human action recognition~\cite{shahroudy2016ntu, liu2019ntu, yan2018spatial, chen2021channel}, 2D keypoints detection~\cite{lin2014microsoft, andriluka14cvpr, sun2019deep, xiao2018simple} and 3D pose estimation~\cite{h36m_pami, reddy2021tessetrack, martinez2017simple},
to dense prediction tasks, such as human parsing \cite{gong2017look, li2017multiple, gong2018instance, chen2014detect} and DensePose prediction~\cite{guler2018densepose}.
Unfortunately, to train a model with reasonable generalizability and robustness, an enormous amount of labeled real data is necessary, which is extremely expensive to collect and annotate.
Therefore, it is desirable to have a versatile pre-train model that can serve as a foundation for all the aforementioned human-centric perception tasks.

With the development of sensors, the human body can be more conveniently perceived and represented in \textbf{multiple modalities}, such as RGB, depth, and infrared.
In this work, we argue that \textit{the multi-modality nature of human-centric data can induce effective representations} that transfer well to various downstream tasks, due to three major \textbf{advantages}:
\textbf{1)} Learning a modal-invariant latent space through pre-training helps efficient task-relevant mutual information extraction. 
\textbf{2)} A single versatile pre-train model on multi-modal data facilitates multiple downstream tasks using various modalities.
\textbf{3)} Our multi-modal pre-train setting bridges heterogeneous human-centric datasets through their common modality, which benefits the generalizability of pre-train models. 

We mainly explore two groups of modalities as shown in Fig. \ref{fig:teaser} a): dense representations (\eg RGB, depth, infrared) and sparse representations (\eg 2D keypoints, 3D pose). 
Dense representations can provide rich texture and/or 3D geometry information. But they are mostly low-level and noisy.
On the contrary, sparse representations obtained by off-the-shelf tools~\cite{8765346, mmpose2020} are semantic and structured. But the sparsity results in insufficient details.
We highlight that it is non-trivial to integrate these heterogeneous modalities into a unified pre-training framework for the following two main \textbf{challenges}: 
\textbf{1)} learning representations suitable for dense prediction tasks in the multi-modality setting;
\textbf{2)} using weak priors from sparse representations effectively for pre-training.

\textbf{Challenge 1: Dense Targets.}
Existing methods~\cite{liu2020p4contrast, hou2021pri3d} perform contrastive learning densely on pixel-level features to achieve view-invariance for dense prediction tasks.
However, those methods require multiple views of a static 3D scene~\cite{dai2017scannet}, which is inapplicable for human-centric applications with only single view.
Furthermore, it is preferable to learn representations that are continuously and orderly distributed over the human body. 
In light of this,
we generalize the widely used InfoNCE~\cite{oord2018representation} and propose a dense intra-sample contrastive learning objective that applies a soft pixel-level contrastive target, which can facilitate learning ordinal and continuous dense feature distributions.

\textbf{Challenge 2: Sparse Priors.}
To employ priors in contrastive learning, previous works~\cite{khosla2020supervised, wei2020can, assran2020supervision} mainly use the supervision to generate semantically positive pairs. However, these methods only focus on the sample-level contrastive learning, which means each sample is encoded to a global embedding.
It is not optimal for human dense prediction tasks.
To this end, we propose a sparse structure-aware contrastive learning target, which uses semantic correspondences across samples as positive pairs to complement positive intra-sample pairs.
Particularly, leveraging sparse human priors leads to an embedding space where semantically corresponding parts are aligned more closely.

To sum up, we propose \textbf{\nickname{}}, a \textbf{H}uman-\textbf{C}entric multi-\textbf{Mo}dal \textbf{Co}ntrastive learning framework for versatile multi-modal pre-training.
To fully leverage multi-modal observations, \nickname{} effectively utilizes both dense measurements and sparse priors using the following three-levels hierarchical contrastive learning objectives:
\textbf{1)} sample-level modality-invariant representation learning; 
\textbf{2)} dense intra-sample contrastive learning; 
\textbf{3)} sparse structure-aware contrastive learning. 
As an effort towards establishing a comprehensive multi-modal human parsing benchmark dataset, we label human segments for RGB-D images from NTU RGB+D dataset~\cite{shahroudy2016ntu}, and contribute the \textbf{NTURGBD-Parsing-4K} dataset.
To evaluate \nickname{}, we transfer our pre-train model to four human-centric downstream tasks using different modalities, including DensePose estimation (RGB)~\cite{guler2018densepose}, human parsing using RGB~\cite{h36m_pami} or depth frames, and 3D pose estimation (depth)~\cite{haque2016towards}.
Under full set and data-efficient training settings,
\nickname{}
constantly achieves better performance than training from scratch or pre-train on ImageNet.
To name a few, as shown in Fig. \ref{fig:teaser} b), we achieve \textbf{7.16\%} improvement in terms of GPS AP on $10\%$ training data of DensePose estimation; 
\textbf{12\%} improvement in terms of mIoU on $20\%$ training data of Human3.6M human parsing.
Moreover, we evaluate the modal-invariance of the latent space learned by \nickname{} for dense prediction on NTURGBD-Parsing-4K with two settings: cross-modality supervision and missing-modality inference.
Compared against conventional contrastive learning targets, our method improves the segmentation mIoU by \textbf{29\%} and \textbf{24\%} for the two settings, respectively.
To the best of our knowledge, we are the first to study multi-modal pre-training for human-centric perception.

The main \textbf{contributions} are summarized below: 
\textbf{1)} As the first endeavor, we provide an in-depth analysis for human-centric pre-training, 
which is formulated as
a challenging multi-modal contrastive learning problem.
\textbf{2)} Together with the novel hierarchical contrastive learning objectives, a comprehensive framework \nickname{} is proposed for effective pre-training for human-centric tasks.
\textbf{3)} Through extensive experiments, \nickname{} achieves superior performance than
existing methods,
and meanwhile shows promising modal-invariance properties.
\textbf{4)} To benefit multi-modal human-centric perception, we contribute an RGB-D human parsing dataset, NTURGBD-Parsing-4K.

\section{Related Work}
\label{sec:relatedwork}

\noindent \textbf{Human-Centric Perception.}
Many efforts have been put into human-centric perception in decades. Lots of work in 2D keypoint detection~\cite{lin2014microsoft, andriluka14cvpr, sun2019deep, xiao2018simple} has achieved robust and accurate performance. 3D pose estimation has long been a challenging problem and is approached from two aspects, lifting from 2D keypoints~\cite{h36m_pami, reddy2021tessetrack, martinez2017simple} and predicting from depth map~\cite{haque2016towards, xiong2019a2j}. Human parsing can be defined in two ways. The first one parses garments together with visible body parts~\cite{gong2017look, li2017multiple, gong2018instance}. The second one only focuses on parsing human parts~\cite{chen2014detect, h36m_pami, hong2021garmentd}. In this work, we focus on the second setting because the depth and 2D keypoints do not contain the texture information needed for garment parsing. There are a few works~\cite{hernandez2012graph, nishi2017generation} about human parsing on depth maps. However, the data and annotations are too coarse or unavailable. To further push the accuracy of human-centric perception, DensePose~\cite{guler2018densepose, tan2021humangps} is proposed to densely model each human body surface point. The cost of DensePose annotation is enormous. Therefore, we also explore data-efficient learning of DensePose.

\noindent \textbf{Multi-Modal Contrastive Learning.}
Multi-modality naturally provides different views of the same sample which fits well into the contrastive learning framework. CMC~\cite{tian2020contrastive} proposes the first multi-view contrastive learning paradigm which takes any number of views. CLIP~\cite{radford2021learning} learns a joint latent space from large-scale paired image-language dataset. Extensive studies~\cite{patrick2021compositions, hazarika2020misa, han2020self, rouditchenko2020avlnet, patrick2020multi, alayrac2020self} focus on video-audio contrastive learning. Recently, 2D-3D contrastive learning~\cite{hou2021pri3d, liu2020p4contrast, liu2021contrastive} has also been studied with the development in 3D computer vision. In this work, aside from commonly used modalities, we also explore the potential of 2D keypoints in human-centric contrastive learning.

\begin{figure}[t]
    \begin{center}
    \includegraphics[width=1.0\linewidth]{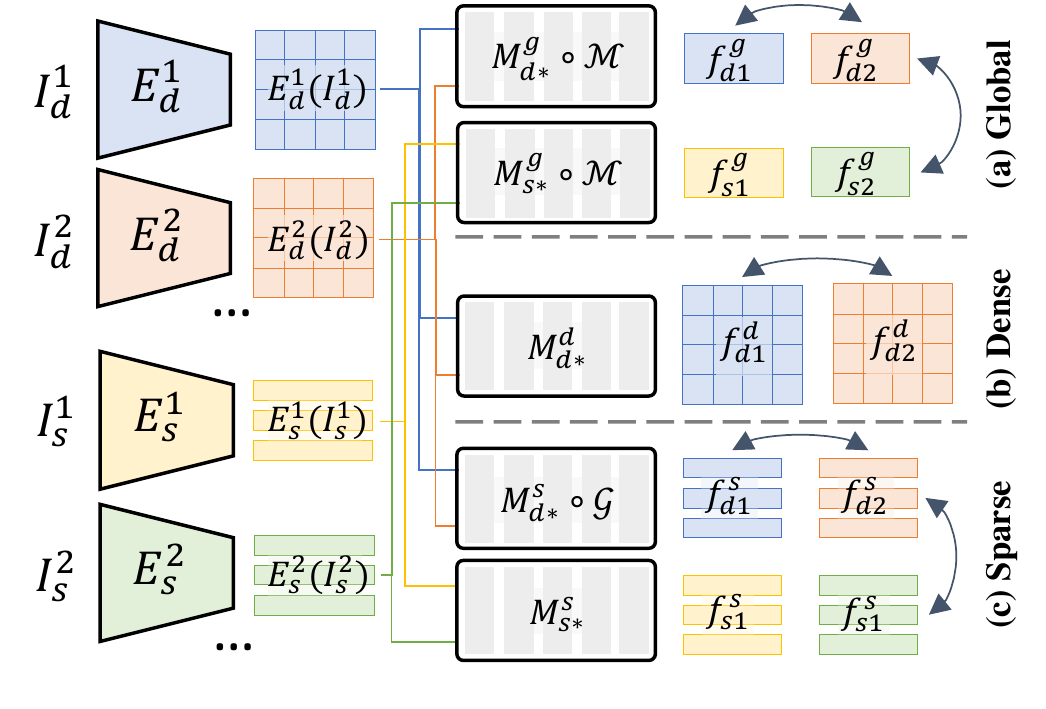}
    \end{center}
    \vspace{-25pt}
    \caption{\textbf{Illustration of the general paradigm of \nickname{}.} We group modalities of human data into dense $I_d^*$ and sparse representations $I_s^*$. Three levels of embeddings are extracted (\ref{sec:general}). Combining the nature of human data and tasks (\ref{sec:int}), we present contrastive learning targets for each level of embedding (\ref{sec:loss}).}
    \label{fig:02_general}
    \vspace{-15pt}
\end{figure}

\section{Our Approach}
\label{sec:method}
In this section, we first introduce the general paradigm of \nickname{} (\ref{sec:general}). Following the design principles (\ref{sec:int}), hierarchical contrastive learning targets are formally introduced (\ref{sec:loss}). Next, an instantiation of \nickname{} is introduced (\ref{sec:inst}). Finally, we propose two applications of \nickname{} to show the versatility (\ref{sec:ext}).

\subsection{\nickname{}}
\label{sec:general}
As shown in Fig. \ref{fig:02_general}, \nickname{} takes multiple modalities of perceived human body as input. The target is to learn human-centric representations, which can be transferred to downstream tasks. The input modalities can be categorized into dense and sparse representations. Dense representations $I_d^*$ are the direct output of imaging sensors, \eg RGB, depth, infrared. They typically contain rich information but are low-level and noisy. Sparse representations are structured abstractions of the human body, \eg 2D keypoints, 3D pose, which can be formulated as graph $I_s^* = G(V,E)$. Different representations of the same view of a human should be spatially aligned, which means intra-sample correspondences can be obtained for dense contrastive learning.
\nickname{} aims to pre-train multiple encoders $E_d^*$ and $E_s^*$ that produce embeddings
of dense representations
and sparse representations for downstream tasks transfer.

To support dense downstream tasks, other than the usual sample-level global embeddings used in \cite{he2020momentum, chen2020simple, chen2020improved, grill2020bootstrap, liu2020self, tian2020contrastive}, we propose to consider different levels of embeddings \ie global embeddings $f^g$, sparse embeddings $f^s$ and dense embeddings  $f^d$
\footnote{For easier understanding of the notations, the superscripts of $f$ and $M$ stand for the kind of embeddings. The subscripts stand for the kind of representations (`g' for `global'; `d' for `dense'; `s' for `sparse').}
, which are defined as follows:
\textbf{1)} For dense representations $I_d$, the global embedding is obtained by applying a mapper network $M^g_d$ to the mean pooling $\mathcal{M}$ of the corresponding feature map, which is formulated as $f_d^g = M^g_d \circ \mathcal{M} \circ E_d(I_d)$. Similarly, for sparse representations $I_s$, the global embedding is defined as $f_s^g = M^g_s \circ \mathcal{M} \circ E_s(I_s)$. 
\textbf{2)} Sparse embeddings have the same size as that of sparse representations. Formally, for sparse representations $I_s = G(V, E)$, where $V \in \mathbb{R}^{J\times K}$, the corresponding sparse embedding is defined as $f_s^s = M^s_s \circ E_s(I_s)$, where $f_s^s \in \mathbb{R}^{J\times K'}$, $M^s_s$ is a mapper network. For dense representations, the corresponding sparse features are pooled from the dense feature map using the correspondences $\mathcal{G}$. Then the sparse features are mapped to sparse embeddings as $f_d^s = M^s_d \circ \mathcal{G} \circ E_d(I_d)$.
\textbf{3)} Dense embeddings are only defined on dense representations, which is formulated as $f_d^d = M^d_d \circ E_d(I_d)$.
With three levels of embeddings defined, we formulate the overall learning objective as
\begin{equation} \label{eq:whole}
    \mathcal{L} = \lambda_g\mathcal{L}_g(f^g) + \lambda_d\mathcal{L}_d(f^d) + \lambda_s\mathcal{L}_s(f^s),
\end{equation}
which is analyzed and explained as follows.

\begin{figure*}[t]
    \begin{center}
        \includegraphics[width=1.0\linewidth]{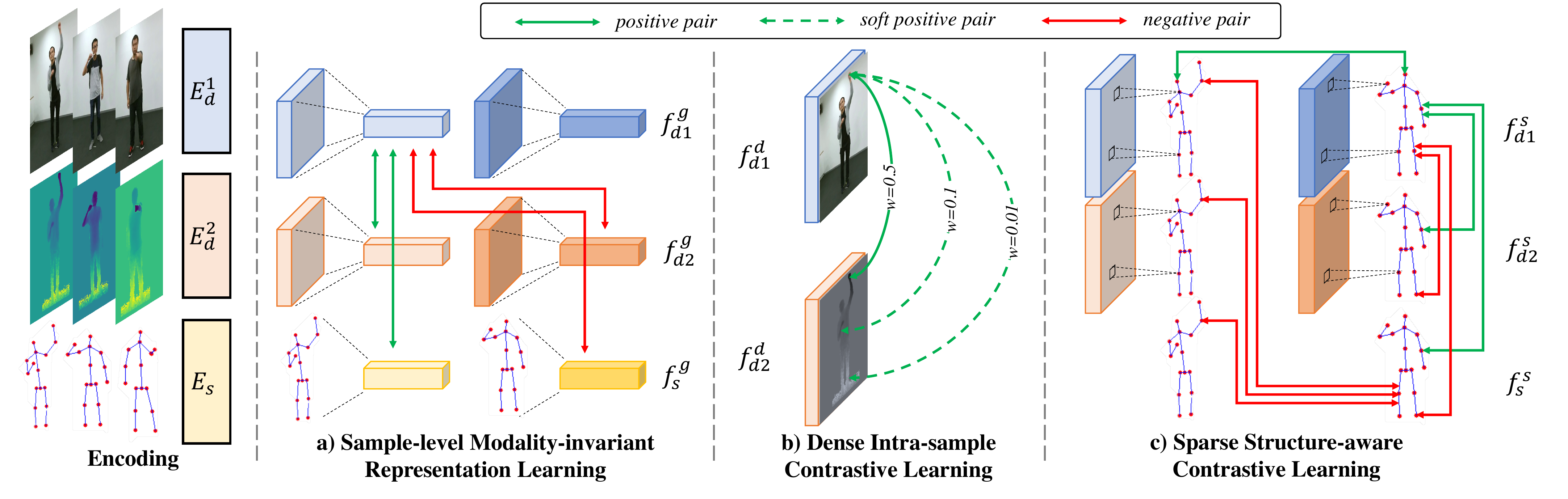}
    \end{center}
    \vspace{-20pt}
    \caption{\textbf{Our Proposed Instantiation of \nickname{}.} For dense representations, we choose to use RGB and depth. For sparse representations, 2D keypoints are used for its convenience to obtain. a) At sample-level, the global embeddings are used for modality-invariant representation learning. b) Between paired dense embeddings, soft contrastive learning target is proposed for continuous and ordinal feature learning. c) Using human prior provided by sparse representations, intra- and inter-sample contrastive learning targets are proposed.}
    \label{fig:03_framework}
    \vspace{-15pt}
\end{figure*}

\subsection{Principles of Learning Targets Design}
\label{sec:int}
In this subsection, we analyze the intuitions when designing learning targets, which makes the following three principles.
\noindent \textbf{1) Mutual Information Maximization}: Inspired by~\cite{wu2018unsupervised, poole2019variational}, we propose to maximize the lower bound on mutual information, which has been proved by many previous works~\cite{he2020momentum, chen2020simple, chen2020improved, tian2020contrastive} to be able to produce strong pre-train models.
\noindent \textbf{2) Continuous and Ordinal Feature Distribution}: Inspired by the property of human-centric perception, it is desirable for the feature maps of the human body to be continuous and ordinal. The human body is a structural and continuous surface. The dense predictions, \eg human parsing~\cite{gong2017look, li2017multiple, gong2018instance}, DensePose~\cite{guler2018densepose}, are also continuous. Therefore, such property should also be reflected in the learned representations. Besides, for an anchor point on human surfaces, closer points have higher probabilities of sharing similar semantics with the anchor point than that of far away points. Therefore, the learned dense representations should also align with such ordinal relationship.
\noindent \textbf{3) Structure-Aware Semantic Consistency}: Sparse representations are abstractions of the human body, which contains valuable structural semantics about the human body. Instead of identity information, the human pose and structure understanding are the keys to our target downstream tasks. Therefore, it is reasonable to eliminate the identity information and enhance the structure information by enforcing structure-aware semantic consistency where semantically close embeddings (\eg embeddings of left hands from different samples) are pulled close and vice versa.

\subsection{Hierarchical Contrastive Learning Targets}
\label{sec:loss}
Based on the above three principles, we formally define hierarchical contrastive learning targets in this subsection.

\noindent \textbf{Sample-level modality-invariant representation learning} aims at learning a joint latent space at the sample level using global embeddings, which fulfills the first principle. Inspired by
\cite{tian2020contrastive}, the learning target can be formulated as
\begin{equation}
    \mathcal{L}_g = -\underset{\underset{f_1^g \in F_1^g}{F_1^g, F_2^g \in S_g}}{\mathbb{E}} \left[ \text{log}\frac{\text{exp}(f_1^g \cdot \bar{f_2^g} / \tau)}{\sum_{f_2^g \in F_2^g} \text{exp}(f_1^g \cdot f_2^g / \tau)} \right],
\end{equation}
\noindent where $F_*^g$ is a set of global embeddings of one modality, $S_g$ is the set of $F_*^g$ of all modalities, $\bar{f_2^g}$ is the embedding of the paired view of that of $f_1^g$, $\tau$ is the temperature. It should be noticed that $f_1^g$ can be sampled from the global embeddings of either dense or sparse representations.

\noindent \textbf{Dense intra-sample contrastive learning} is operated on the paired dense representations. For any two paired dense embeddings $f_{d1}^d, f_{d2}^d \in \mathbb{R}^{H\times W\times K'}$,
to simultaneously satisfy the first and the second principle, the dense intra-sample contrastive learning target between them is defined in a `soft' way as
\begin{equation}
    \mathcal{L}_d^{12} = -\underset{\underset{m, n}{x, y}}{\mathbb{E}} \left[ \mathcal{W}_{xy}^{mn} \text{log} \frac{\text{exp}(f_{d1}^d(x, y) \cdot f_{d2}^d(m, n)) / \tau)}{\sum\limits_{x', y'} \text{exp}(f_{d1}^d(x,y) \cdot f_{d2}^d(x', y') / \tau)} \right],
\end{equation}
\noindent where $\mathcal{W}_{xy}^{mn}$ is the weight, $\tau$ is the temperature, $(x, y), (m, n), (x', y')$ are coordinates on the dense representation, $1 \le x, x', m \le H$, $1 \le y, y', n \le W$. The above equation is a generalized version of InfoNCE~\cite{oord2018representation}. InfoNCE is a special case when $\mathcal{W}_{xy}^{mn}$ is set to $1$ if $x = m$ and $y = n$ else $0$.
We use the normalized distances as the weights, which is formulated as 
\begin{equation}
    \mathcal{W}_{xy}^{mn} = \frac{\text{exp}(\sqrt{(x - m)^2 + (y - n)^2})}{\sum_{x', y'}\text{exp}(\sqrt{(x - x')^2 + (y - y')^2})}.
\end{equation}
For each pair of dense representations, the above learning target is calculated between each pair of dense embeddings. Therefore, the whole learning target is defined as
\begin{equation}
    \mathcal{L}_d = \underset{\underset{f_{d1}^d, f_{d2}^d \in F_1^d, F_2^d}{F_1^d, F_2^d \in S_d}}{\mathbb{E}} \mathcal{L}_d^{12}(f_{d1}^d, f_{d2}^d),
\end{equation}
where $F_*^d$ is a set of dense embeddings of one modality, $S_d$ is the set of all $F_*^d$, $f_{d1}^d$ and $f_{d2}^d$ are two paired embeddings. It should be noticed that the `soft' learning target cannot guarantee an ordinal feature distribution. Instead, it serves as a computationally efficient relaxation of the requirement of ordinal distribution.

\noindent \textbf{Sparse structure-aware contrastive learning} takes two sparse representations $f_1^s$ and $f_2^s$ as inputs. The paired features $f_{1j}^s$ and $f_{2j}^s$ (\ie features of the $j$-th joint) should be pulled close while unpaired features are pushed away. The two sparse representations can be sampled from the same or different modalities, intra- or inter-sample. The intra-sample alignment satisfies the first principle. The inter-sample alignment follows the third principle. The sparse structure-aware contrastive learning target is formulated as
\begin{equation}
    \mathcal{L}_s = -\underset{\underset{j; f_1^s, f_2^s \in \{F_1^s, F_2^s\}}{F_1^s, F_2^s \in S_s}}{\mathbb{E}} \left[\text{log}\frac{\text{exp}(f_{1j}^{s}\cdot f_{2j}^{s} / \tau)}{\sum\limits_{j'; f_i^s \in \{F_1^s, F_2^s\}} \text{exp}(f_{1j}^s \cdot f_{ij'}^s / \tau)} \right],
\end{equation}
where $F_*^s$ is a set of sparse embeddings of one modality, $S_s$ is the set of $F_*^s$, $\tau$ is the temperature, $f_1^s, f_2^s$ are sampled from the union of $F_1^s$ and $F_2^s$.
To conclude, the overall learning target is formulated as Eq. \ref{eq:whole},
where $\lambda_*$ are the weights to balance the targets.

\subsection{Instantiation of \nickname{}}
\label{sec:inst}
In this section, we introduce an instantiation of \nickname{}. As shown in Fig. \ref{fig:03_framework}, for dense representations, we use RGB and depth. Large-scale paired human RGB and depth data is easy to obtain with affordable sensors \eg Kinect. These two modalities are the most commonly encountered in human-centric tasks~\cite{chen2014detect, h36m_pami, gong2017look, li2017multiple, gong2018instance}. Moreover, a proper pre-train model for depth is highly desired. Therefore, RGB and depth are reasonable choices of human dense representations, both of which are easy to acquire and important to downstream tasks. For sparse representations, 2D keypoints are used, which provide positions of human body joints in the image coordinate. Off-the-shelf tools~\cite{8765346, mmpose2020} are available to quickly and robustly extract human 2D keypoints given RGB images. Using 2D keypoints as the sparse representation is a good balance between the amount of human prior and acquisition difficulty.

For RGB inputs $I_d^1$, an image encoder $E_d^1$~\cite{sun2019deep} is applied to obtain feature maps $E_d^1(I_d^1)$. Similarly, for depth inputs $I_d^2$, an image encoder~\cite{sun2019deep} or 3D encoder~\cite{qi2017pointnet, qi2017pointnet++} $E_d^2$ can be applied to extract feature maps $E_d^2(I_d^2)$. 2D keypoints $I_s$ are encoded by a GCN-based encoder~\cite{zhao2019semantic} $E_s$ to produce sparse features $E_s(I_s)$. Mapper networks comprise a single linear layer and a normalization operation.

As for the implementation of contrastive learning targets, we choose to use a memory pool to store all the global embeddings which are updated in a momentum way. Sparse and dense embeddings cannot all fit in memory. Therefore, for the last two types of contrastive learning targets, the negative samples are sampled within a mini-batch.

\begin{figure}[t]
    \begin{center}
    \includegraphics[width=1.0\linewidth]{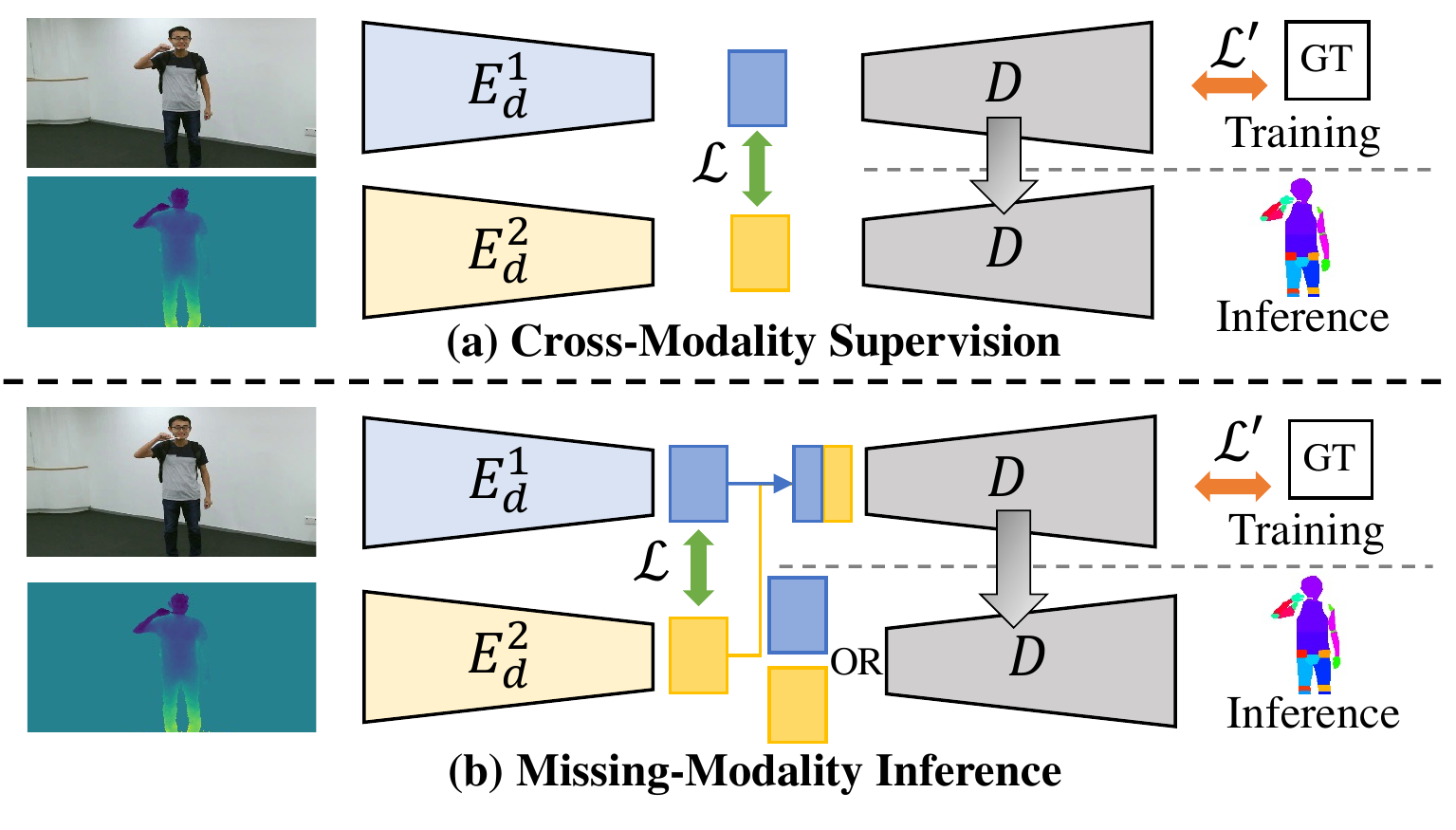}
    \end{center}
    \vspace{-20pt}
    \caption{\textbf{Pipelines of Two Applications of \nickname{}.}}
    \label{fig:04_otherapp}
    \vspace{-15pt}
\end{figure}

\subsection{Versatility of \nickname{}}
\label{sec:ext}
On top of the pre-train framework \nickname{}, we propose to further extend it on two direct applications: cross-modality supervision and missing-modality inference.
The extensions are based on the key design of \nickname{}: dense intra-sample contrastive learning target. With the feature maps of different modalities aligned, it is straightforward to implement the two extensions, which are shown in Fig. \ref{fig:04_otherapp}.

\noindent\textbf{Cross-Modality Supervision} is a novel task where we train the network on the source modality, while test on the target modality. This is a practical scenario where people transfer the knowledge of some single modality dataset to other modalities. 
At training time, an additional downstream task head (\eg segmentation head) $D$ is attached to the backbone of the source modality. The hierarchical contrastive learning targets $\mathcal{L}$ together with downstream task loss $\mathcal{L}'$ are used for end-to-end training. At inference time, $D$ is attached to the backbone of the target modality. The extracted feature maps of the target modality are passed to $D$ for prediction.

\noindent\textbf{Missing-Modality Inference} is another novel task where we train the network using multi-modal data and inference on single modality. Multi-modal data collection in practice would inevitably result in data with incomplete modalities, which brings the requirement of missing-modality inference.
At training time, the feature maps of multiple modalities are fused using max-pooling and fed to a downstream task head $D$. Similarly, hierarchical contrastive learning targets $\mathcal{L}$ and downstream task loss $\mathcal{L}'$ are used for co-training. At inference time, the feature map of a single modality is passed to $D$ for missing-modality inference.

\begin{figure}[t]
    \begin{center}
    \includegraphics[width=1.0\linewidth]{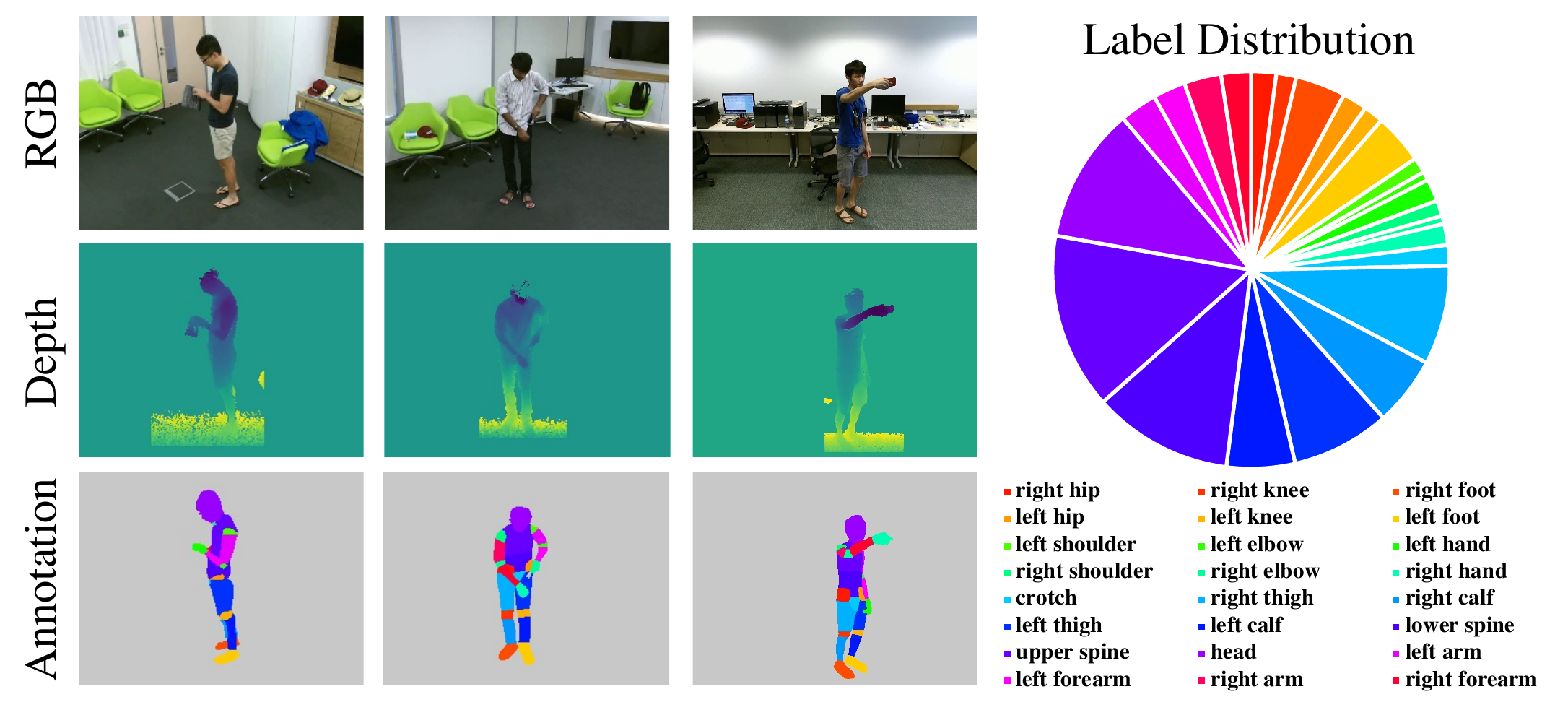}
    \end{center}
    \vspace{-20pt}
    \caption{Illustration of the RGB-D human parsing dataset \textbf{NTURGBD-Parsing-4K}.}
    \label{fig:dataset}
    \vspace{-15pt}
\end{figure}


\begin{table*}[ht]
    \caption{DensePose Estimation Results on COCO. \textsuperscript{*} randomly initializes the model before pre-training. $^\dagger$ initializes the model by ImageNet pre-train before pre-training. All results in [\%].}
    \vspace{-15pt}
    \begin{center}
    \small{
        \addtolength{\tabcolsep}{-2pt}
        \begin{tabular}{l|c|cccc|cccc}
            \Xhline{1pt}
            \multirow{2}{*}{Method} & \multirow{2}{*}{Pre-train Datasets} & \multicolumn{4}{c|}{Full Data} & \multicolumn{4}{c}{10\% Data} \\
            & & BBox AP & GPS AP & GPSM AP & IOU AP & BBox AP & GPS AP & GPSM AP & IoU AP \\
            \hline\hline
            From Scratch           & -            & 57.27 & 62.03 & 63.61 & 65.88 & 39.38 & 35.75 & 41.62 & 49.92 \\
            CMC\textsuperscript{*}~\cite{tian2020contrastive} & NTURGBD+MPII & 60.33 & 64.97 & 65.66 & 66.96 & 44.92 & 43.84 & 47.94 & 54.00 \\
            MMV\textsuperscript{*}~\cite{alayrac2020self} & NTURGBD+MPII & 59.89 & 64.23 & 65.47 & 67.03 & 43.24 & 41.40 & 45.99 & 52.52 \\
            Ours\textsuperscript{*}& NTURGBD+MPII & \textbf{61.33} & \textbf{65.89} & \textbf{66.92} & \textbf{67.66} & \textbf{47.76} & \textbf{48.47} & \textbf{51.65} & \textbf{56.15} \\
            \hline\hline
            IN Pre-train            & -            & 62.66 & 66.48 & 67.42 & 68.63 & 48.28 & 44.34 & 49.11 & 56.11 \\
            CMC$^\dagger$~\cite{tian2020contrastive}          & NTURGBD+MPII & 62.76 & 66.16 & 67.30 & 68.06 & 49.21 & 48.82 & 52.57 & 57.94 \\
            MMV$^\dagger$~\cite{alayrac2020self}          & NTURGBD+MPII & 62.97 & 66.67 & 67.51 & 68.29 & 50.16 & 50.28 & 53.54 & 58.32 \\
            Ours$^\dagger$         & NTURGBD+MPII & \textbf{63.11} & \textbf{67.33} & \textbf{68.12} & \textbf{68.72} & \textbf{50.29} & \textbf{51.50} & \textbf{54.47} & \textbf{58.66} \\
            \hline\hline
            CMC$^\dagger$~\cite{tian2020contrastive}         & NTURGBD+COCO & \textbf{63.58} & 67.22 & 67.77 & 68.46 & 51.77 & 53.53 & 56.18 & 59.37 \\
            Ours$^\dagger$        & NTURGBD+COCO & 62.95 & \textbf{67.77} & \textbf{68.29} & \textbf{68.63} & \textbf{52.18} & \textbf{54.01} & \textbf{56.64} & \textbf{59.93} \\
            \Xhline{1pt}
        \end{tabular}
    }
    \end{center}
    \label{tab:densepose}
    \vspace{-15pt}
\end{table*}

\begin{table*}[ht]
    \caption{Human Parsing Results on Human3.6M. \textsuperscript{*} randomly initializes the model before pre-training. $^\dagger$ initializes the model by ImageNet pre-train before pre-training. All results in [\%].}
    \vspace{-15pt}
    \begin{center}
    \small{
        \begin{tabular}{l|ccc|ccc|ccc|ccc}
            \Xhline{1pt}
            \multirow{2}{*}{Method} & \multicolumn{3}{c|}{Full Data} & \multicolumn{3}{c|}{20\% Data} & \multicolumn{3}{c|}{10\% Data} & \multicolumn{3}{c}{1\% Data} \\
            & mIoU & mAcc & aAcc & mIoU & mAcc & aAcc & mIoU & mAcc & aAcc & mIoU & mAcc & aAcc \\
            \hline\hline
            From Scratch           & 44.13 & 58.88 & 98.82 & 42.41 & 56.25 & 98.81 & 32.61 & 43.76 & 98.52 & 7.27 & 10.97 & 97.45 \\
            CMC\textsuperscript{*}~\cite{tian2020contrastive} & 54.33 & 68.01 & 99.09 & 52.10 & 65.65 & 99.03 & 48.37 & 61.18 & 98.95 & 14.61 & 20.07 & 98.07 \\
            MMV\textsuperscript{*}~\cite{alayrac2020self} & 52.69 & 65.82 & 99.06 & 50.66 & 63.55 & 99.01 & 46.23 & 58.52 & 98.90 & 12.86 & 17.10 & 97.94 \\
            Ours\textsuperscript{*}& \textbf{61.36} & \textbf{75.09} & \textbf{99.25} & \textbf{59.17} & \textbf{73.44} & \textbf{99.19} & \textbf{57.08} & \textbf{71.75} & \textbf{99.13} & \textbf{16.55} & \textbf{22.27} & \textbf{98.18} \\
            \hline\hline
            IN Pre-train              & 56.90 & 69.94 & 99.14 & 48.86 & 60.75 & 98.97 & 44.55 & 56.86 & 98.87 & 14.65 & 20.22 & 98.09 \\
            CMC$^\dagger$~\cite{tian2020contrastive}            & 58.93 & 71.70 & 99.20 & 57.41 & 70.13 & 99.17 & 54.35 & 67.47 & 99.09 & 17.77 & 23.77 & 98.20 \\
            MMV$^\dagger$~\cite{alayrac2020self}            & 59.08 & 71.57 & 99.20 & 57.28 & 69.69 & 99.17 & 53.86 & 66.46 & 99.08 & 17.66 & 23.54 & 98.20 \\
            Ours$^\dagger$           & \textbf{62.50} & \textbf{75.84} & \textbf{99.27} & \textbf{60.85} & \textbf{74.23} & \textbf{99.23} & \textbf{58.28} & \textbf{71.99} & \textbf{99.17} & \textbf{20.78} & \textbf{27.52} & \textbf{98.34} \\
            \Xhline{1pt}
        \end{tabular}
    }
    \end{center}
    \label{tab:human36m}
    \vspace{-20pt}
\end{table*}

\vspace{10pt}

\section{NTURGBD-Parsing-4K Dataset}
Although RGB human parsing has been well studied~\cite{chen2014detect, gong2017look, li2017multiple, gong2018instance}, human parsing on depth~\cite{hernandez2012graph, nishi2017generation} or RGB-D data has not been fully addressed due to the lack of labeled data. Therefore, we contribute the first RGB-D human parsing dataset: NTURGBD-Parsing-4K. The RGB and depth are uniformly sampled from NTU RGB+D (60/120)~\cite{shahroudy2016ntu, liu2019ntu}. As shown in Fig. \ref{fig:dataset}, we annotate 24 human parts for paired RGB-D data. The partition protocols follow that of \cite{h36m_pami}. The train and test set both have $1963$ samples. The whole dataset contains $3926$ samples. Hopefully, by contributing this dataset, we could promote the development of both human perception and multi-modality learning.

\section{Experiments}
\label{sec:exp}

\subsection{Experimental Setup}
\noindent \textbf{Implementation Details.}
The default RGB and depth encoders are HRNet-W18~\cite{sun2019deep}.
The default datasets for pre-train are NTU RGB+D~\cite{liu2019ntu} and MPII~\cite{andriluka14cvpr}. The former provides paired indoor human RGB, depth, and 2D keypoints, The latter provides in-the-wild human RGB and 2D keypoints. Mixing human data from different domains helps our pre-train models adapt to a wilder domain.

\noindent \textbf{Downstream Tasks.}
We test our pre-train models on four different human-centric downstream tasks, two on RGB images and two on depth.
\textbf{1)} DensePose estimation on COCO~\cite{guler2018densepose}: DensePose aims at mapping pixels of the observed human body to the surface of a 3D human body, which is a highly challenging task.
\textbf{2)} RGB human parsing on Human3.6M~\cite{h36m_pami}.
Human3.6M provides pure human part segmentation, which aligns with our objectives. 
We uniformly sample 2fps of the video for training and evaluation.
\textbf{3)} Depth human parsing on NTURGBD-Parsing-4K.
\textbf{4)} 3D pose estimation from depth maps on ITOP~\cite{haque2016towards} (only side view).
For all the above downstream tasks, we use the pre-train backbones for end-to-end fine-tune.

\noindent \textbf{Comparison Methods.}
Since there are few previous human-centric multi-modal pre-train methods, we propose to use general multi-modal contrastive learning methods CMC~\cite{tian2020contrastive} and MMV~\cite{alayrac2020self} as the baselines. Although there are other multi-modal contrastive learning works, they either require the multi-view calibration~\cite{hou2021pri3d} or focus on multi-modal downstream tasks~\cite{liu2021contrastive, hazarika2020misa} and therefore are not suitable for comparison. In addition, for RGB tasks, we also experiment under two settings, one initializes encoders with supervised ImageNet~\cite{krizhevsky2012imagenet} (IN) pre-train while the other does not.

\begin{table*}[ht]
    \caption{Ablation Study on Densepose/ Human3.6M/ ITOP/ NTURGBD-Parsing-4K. All results in [\%].}
    \vspace{-15pt}
    \begin{center}
    \small{
        \begin{tabular}{l|cccc|cc|cc|cc}
            \Xhline{1pt}
            \multirow{2}{*}{Method} & \multicolumn{4}{c|}{DensePose $10\%$} & \multicolumn{2}{c|}{ITOP $0.1\%$/ $0.2\%$} & \multicolumn{2}{c|}{Human3.6M $10\%$} & \multicolumn{2}{c}{NTURGBD $20\%$} \\
            & BBox & GPS & GPSM & \multicolumn{1}{c|}{IoU} & Acc & \multicolumn{1}{c|}{Acc} & mIoU & \multicolumn{1}{c|}{mAcc} & mIoU & mAcc \\
            \hline\hline
            Sample-level Mod-invariant  & 49.21 & 48.82 & 52.57 & 57.94 & 57.73 & 50.08 & 54.35 & 67.47 & 30.40 & 51.54 \\
            $+$ Hard Dense Intra-sample & 49.40 & 49.14 & 52.49 & 57.30 & 56.43 & 54.05 & 55.36 & 68.43 & 31.26 & 51.54 \\
            $+$ Soft Dense Intra-sample & 50.21 & 50.25 & 53.42 & 57.70 & 62.33 & 51.50 & 56.35 & 69.26 & 32.20 & 51.06 \\
            $+$ Sparse Structure-aware  & \textbf{50.29} & \textbf{51.50} & \textbf{54.47} & \textbf{58.66} & \textbf{65.83} & \textbf{62.36} & \textbf{58.28} & \textbf{71.99} & \textbf{35.01} & \textbf{52.55} \\
            \Xhline{1pt}
        \end{tabular}
    }
    \end{center}
    \label{tab:ablation}
    \vspace{-25pt}
\end{table*}

\subsection{Performance on Downstream Tasks}
\noindent \textbf{DensePose Estimation.}
As shown in Tab. \ref{tab:densepose}, we test DensePose estimation~\cite{guler2018densepose} under two settings: full and $10\%$ of the training data. The trained models are tested on the full validation set of DensePose. Firstly, if not using IN pre-train, our pre-train model significantly outperforms both `From Scratch' and two baseline methods. Especially under $10\%$ of training data, \textbf{12.7\%} improvement in terms of GPS AP is observed. And our pre-train model even outperforms that using IN pre-train by \textbf{4.13\%} in terms of GPS AP. When we use IN pre-train as initialization, which is a common practice for 2D tasks, our method still outperforms all the baselines. Our method surpasses IN pre-train by \textbf{7.2\%} and \textbf{5.4\%} in terms of GPS/GPSM AP under $10\%$ setting. To further test the performance of in-domain transfer, we also pre-train models using training sets of NTU RGB+D and COCO. The performance gain under $10\%$ setting further improves to \textbf{9.7\%} and \textbf{7.5\%} in terms of GPS/GPSM AP.

\noindent \textbf{RGB Human Parsing.}
As shown in Tab. \ref{tab:human36m}, we test four settings on Human3.6M~\cite{h36m_pami}: full, $20\%$, $10\%$ and $1\%$ training data. In all settings, our method outperforms all baselines in all metrics. On full training data, we outperform IN pre-train by \textbf{5.6\%} in terms of mIoU. The performance gain increases with the amount of training data decreases. It is worth noticing that with only $10\%$ of training data, our method outperforms IN pre-train with full training data.

\begin{table}[ht]
    \vspace{-5pt}
    \caption{Human Parsing Results on NTURGBD-Parsing-4K [\%].}
    \vspace{-15pt}
    \begin{center}
    \small{
        \addtolength{\tabcolsep}{-2pt}
        \begin{tabular}{l|ccc|ccc}
            \Xhline{1pt}
            \multirow{2}{*}{Method} & \multicolumn{3}{c|}{Full Data} & \multicolumn{3}{c}{20\% Data} \\
            & mIoU & mAcc & aAcc & mIoU & mAcc & aAcc \\
            \hline\hline
            IN Pre-train             & 37.49 & 57.52 & 98.36 & 28.56 & 46.81 & 98.10 \\
            CMC~\cite{tian2020contrastive}                     & 38.20 & 58.73 & 98.39 & 30.40 & 51.54 & 98.02 \\
            MMV~\cite{alayrac2020self}                     & 38.09 & 58.49 & 98.37 & 30.41 & 50.62 & 98.07 \\
            Ours                    & \textbf{39.32} & \textbf{58.79} & \textbf{98.47} & \textbf{35.01} & \textbf{52.55} & \textbf{98.53} \\
            \Xhline{1pt}
        \end{tabular}
    }
    \end{center}
    \label{tab:nturgbd}
    \vspace{-15pt}
\end{table}
\noindent \textbf{Depth Human Parsing.}
As shown in Tab. \ref{tab:nturgbd}, we test the pre-train depth backbone on our proposed Dataset NTURGBD-Parsing-4K with all training data and $20\%$ training data. We outperform all baselines on two settings. Especially, only using $20\%$ of training data, we surpass IN pre-train by \textbf{6.4\%} and MMV~\cite{alayrac2020self} by \textbf{4.6\%} in terms of mIoU.

\begin{table}[ht]
    \vspace{-5pt}
    \caption{3D Pose Estimation Results on ITOP. All results in [\%].}
    \vspace{-15pt}
    \begin{center}
    \small{
        \addtolength{\tabcolsep}{-1.5pt}
        \begin{tabular}{l|cccccc}
            \Xhline{1pt}
            Method & 100\% & 10\% & 1\% & 0.5\% & 0.2\% & 0.1\% \\
            \hline\hline
            IN Pre-train            & 85.19 & 83.44 & 77.20 & 54.31 & 13.27 & 14.21 \\
            CMC~\cite{tian2020contrastive}                    & 87.08 & 85.36 & 79.49 & 75.07 & 57.73 & 50.08 \\
            MMV~\cite{alayrac2020self}                    & 86.13 & 83.49 & \textbf{79.70} & 71.70 & 60.83 & 54.44 \\
            Ours                   & \textbf{87.19} & \textbf{85.49} & 78.71 & \textbf{76.34} & \textbf{65.83} & \textbf{62.36} \\
            \Xhline{1pt}
        \end{tabular}
    }
    \end{center}
    \label{tab:itop}
    \vspace{-15pt}
\end{table}
\noindent \textbf{3D Pose Estimation.}
As shown in Tab. \ref{tab:itop}, we test the pre-train depth backbone on ITOP~\cite{haque2016towards} with six different ratios of training data. Our pre-train model outperforms all baselines on most settings. With only $10\%$ training data, the accuracy of our method outperforms that of IN pre-train with all training data. It is also worth noticing that $0.1\%$ of training data are $17$ samples, which makes this a few-shot learning setting. With such limited training data, IN pre-train barely produce meaningful results, while our method improves the accuracy by \textbf{48.2\%}.

\subsection{Ablation Study}
In this subsection, we perform a thorough ablation study on \nickname{} to justify the design choices. As shown in Tab. \ref{tab:ablation}, we firstly report the results of only applying sample-level modality-invariant representation learning. Then we add dense intra-sample contrastive learning and sparse structure-aware contrastive learning in order. To further demonstrate the effect of the `soft' design in dense intra-sample contrastive learning, we also report results of the `hard' learning target, which takes the form of a classic InfoNCE~\cite{oord2018representation}. We report the results of the ablation study on all four downstream tasks under data-efficient settings.

For DensePose estimation, it is important to learn feature maps that are continuously and ordinally distributed, which is the expected result of soft dense intra-sample contrastive learning. The performance gain of the soft learning target over the hard counterpart justifies the observation and the learning target design. The dense intra-sample contrastive learning also shows superiority on three other downstream tasks, which shows the importance of fine-grained contrastive learning targets for dense prediction tasks.

Explicitly injecting human prior into the network through sparse structure-aware contrastive learning also proves its effectiveness by further improving the performance on DensePose. 
Thanks to the strong hints provided by 2D keypoints, the performance of 3D pose estimation is improved. Moreover, the sparse structure-aware contrastive learning boosts the performance of human parsing both on RGB and depth maps by \textbf{1.9\%} and \textbf{2.8\%} respectively in terms of mIoU. Although 2D keypoints are sparse priors, they still provide the rough location of each part of the human body,
which facilitate the feature alignment of same body parts.
To summarize, the sparse and dense learning targets both contribute to the performance of our methods, which is in line with our analysis.

\subsection{Performance on \nickname{} Versatility}
\noindent \textbf{Cross-Modality Supervision.} 
We test the cross-modality supervision pipeline on the task of human parsing on NTURGBD-Parsing-4K because it has two modalities and respective dense annotations. Two baseline methods are adopted: 1) using CMC~\cite{tian2020contrastive} contrastive learning target; 2) no contrastive learning target. For a fair comparison, the backbones of all methods are initialized by CMC~\cite{tian2020contrastive} pre-train. At training time, the target modality of training data is not available. We experiment on two settings where we supervise on RGB, test on depth (RGB $\rightarrow$ Depth), and vice versa (Depth $\rightarrow$ RGB). As shown in Tab. \ref{tab:crossmod}, our method outperforms both baselines under two settings. Specifically, our method improves the mIoU of both settings by \textbf{29.2\%} and \textbf{23.0\%}, respectively. Even compared to methods with direct supervision, we can achieve comparable results.

\begin{table}[t]
    \caption{Cross-Modality Supervised Human Parsing Results on NTURGBD-Parsing-4K. All results in [\%].}
    \vspace{-15pt}
    \begin{center}
    \small{
        \addtolength{\tabcolsep}{-2pt}
        \begin{tabular}{l|ccc|ccc}
            \Xhline{1pt}
            \multirow{2}{*}{Method} & \multicolumn{3}{c|}{RGB $\rightarrow$ Depth} & \multicolumn{3}{c}{Depth $\rightarrow$ RGB} \\
            & mIoU & mAcc & aAcc & mIoU & mAcc & aAcc \\
            \hline \hline
            No Contrastive & 3.94  & 4.36  & 92.24 & 3.71 & 4.03 & 91.63 \\
            CMC~\cite{tian2020contrastive}            & 3.86  & 5.59  & 86.81 & 3.85 & 4.27 & 91.75 \\
            Ours           & \textbf{33.19} & \textbf{54.38} & \textbf{94.70} & \textbf{26.80} & \textbf{48.80} & \textbf{92.84} \\
            \Xhline{1pt}
        \end{tabular}
    }
    \end{center}
    \label{tab:crossmod}
    \vspace{-20pt}
\end{table}


\begin{table}[ht]
    \vspace{-5pt}
    \caption{Missing-Modality Human Parsing Results on NTURGBD-Depth. All results in [\%].}
    \vspace{-15pt}
    \begin{center}
    \small{
        \addtolength{\tabcolsep}{-2pt}
        \begin{tabular}{l|ccc|ccc}
            \Xhline{1pt}
            \multirow{2}{*}{Method} & \multicolumn{3}{c|}{Only RGB} & \multicolumn{3}{c}{Only Depth} \\
            & mIoU & mAcc & aAcc & mIoU & mAcc & aAcc \\
            \hline \hline
            No Contrastive & 13.45 & 14.77 & 93.35 & 24.41 & 30.49 & 95.27 \\
            CMC~\cite{tian2020contrastive}            & 19.62 & 28.19 & 92.94 & 16.58 & 19.83 & 93.94 \\
            Ours           & \textbf{43.88} & \textbf{64.27} & \textbf{96.15} & \textbf{43.98} & \textbf{63.66} & \textbf{96.34} \\
            \Xhline{1pt}
        \end{tabular}
    }
    \end{center}
    \label{tab:missingmod}
    \vspace{-15pt}
\end{table}
\noindent \textbf{Missing-Modality Inference.}
For missing-modality inference, we report the experiments on the same dataset and same baselines as above. As shown in Tab. \ref{tab:missingmod}, with no pixel-level alignment, the two baseline methods struggle in two missing-modality settings \ie `Only RGB' and `Only Depth'. While our method improves the segmentation mIoU by \textbf{24.3\%} and \textbf{19.6\%} on two settings.

\begin{figure}[ht]
    \begin{center}
    \includegraphics[width=1.0\linewidth]{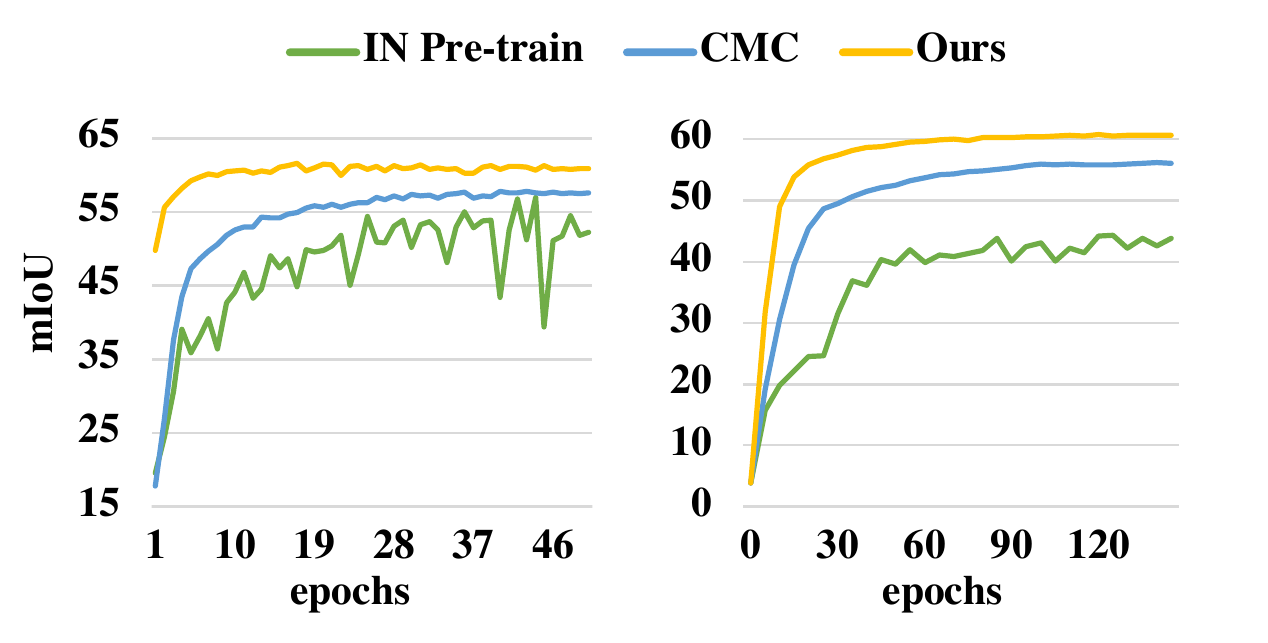}
    \end{center}
    \vspace{-20pt}
    \caption{Validation mIoU Changes with Training Epochs Increase. Left: Human3.6M human parsing full training set. Right: Human3.6M human parsing 20\% training set.}
    \label{fig:fast_conv}
    \vspace{-10pt}
\end{figure}

\subsection{Further Analysis}
\noindent \textbf{Faster Convergence.}
One of the advantages of pre-training is the fast convergence speed when transferred to downstream tasks. Our \nickname{} also shows superiority in this feature. We log the validation mIoU of Human3.6M human parsing at different training epochs. As shown in Fig. \ref{fig:fast_conv}, compared with IN pre-train and CMC~\cite{tian2020contrastive}, our pre-train model is able to converge within a few training epochs in both the full training data and data-efficient settings.

\begin{table}[t]
    \caption{Experiments on changing the backbone. * stands for `IN Pre-train' for DensePose and `From Scratch' for NTURGBD-Parsing-4K. All results in [\%].}
    \vspace{-15pt}
    \begin{center}
    \small{
        \addtolength{\tabcolsep}{-2pt}
        \begin{tabular}{l|cccc|cc}
            \Xhline{1pt}
            \multirow{2}{*}{Method} & \multicolumn{4}{c|}{DensePose $10\%$} & \multicolumn{2}{c}{NTURGBD $20\%$} \\
            & BBox & GPS & GPSM & IoU & mIoU & mAcc \\
            \hline \hline
            *    & \textbf{55.10} & 54.60 & 57.60 & 61.73 & 45.36 & 59.51 \\
            CMC~\cite{tian2020contrastive}  & 53.88 & 54.62 & 57.46 & 61.14 & 48.74 & 62.94 \\
            Ours & 54.55 & \textbf{55.80} & \textbf{58.36} & \textbf{61.75} & \textbf{49.43} & \textbf{63.52} \\
            \Xhline{1pt}
        \end{tabular}
    }
    \end{center}
    \label{tab:changebackbone}
    \vspace{-16pt}
\end{table}
\noindent \textbf{Changing Backbone.}
So far our experiments are all performed on HRNet-W18. To further demonstrate \nickname{}'s performance on other backbones, for the 2D backbone, we also experiment with HRNet-W32~\cite{sun2019deep}. For the depth backbone, we choose to test with PointNet++~\cite{qi2017pointnet++}. For the RGB pre-train model, we experiment on the $10\%$ DensePose estimation. For the depth pre-train model, we experiment on the $20\%$ NTURGBD-Parsing-4K. As shown in Tab. \ref{tab:changebackbone}, our method outperforms its pre-train counterparts by a reasonable margin, which is in line with our previous experimental results.

\section{Discussion and Conclusion}
In this work, we propose the first versatile multi-modal pre-training framework \nickname{} specifically designed for human-centric perception tasks. Hierarchical contrastive learning targets are designed based on the nature of human datasets and the requirements of human-centric downstream tasks. Extensive experiments on four different human downstream tasks of different modalities demonstrated the effectiveness of our pre-training framework. We contribute a new RGB-D human parsing dataset NTURGBD-Parsing-4K to support research of human perception on RGB-D data. Besides downstream task transfer, we also propose two novel applications of \nickname{} to show its versatility and ability in cross-modal reasoning.

\noindent \textbf{Potential Negative Impacts \& Limitations.}
Usage of large amounts of data and long training time might negatively impact the environment. Moreover, even though we did not collect any new human data in this work, human data collection could happen if our framework is used in other applications, which potentially raises privacy concerns. As for the limitations, due to limited resources, we could only experiment with one possible instantiation of \nickname{}. And for the same reason, even though the theoretical possibility exists, we do not have the chance to further scale up the amount of human dataset and network size.

\noindent \textbf{Acknowledgments} \quad
This work is supported by NTU NAP, MOE AcRF Tier 2 (T2EP20221-0033), and under the RIE2020 Industry Alignment Fund – Industry Collaboration Projects (IAF-ICP) Funding Initiative, as well as cash and in-kind contribution from the industry partner(s).

\clearpage

{\small
\bibliographystyle{ieee_fullname}
\bibliography{egbib}
}

\clearpage

\setcounter{section}{0}

\begin{center}
    {\large \textbf{Supplementary Material}}
\end{center}

In this supplementary material, we provide more implementation details of \nickname{}, downstream tasks and two applications (Sec. \ref{sec:impl}). More detailed quantitative and qualitative results of downstream tasks are also illustrated (Sec. \ref{sec:moreres} and Sec. \ref{sec:vis}).

\section{Implementation Details} \label{sec:impl}

\subsection{\nickname{}}

\noindent\textbf{Network Details.} For the proposed instantiation of \nickname{}, we implement the sample-level modality-invariant representation learning target by maintaining a memory pool, which is adapted from an open-sourced implementation \footnote{\url{https://github.com/HobbitLong/PyContrast}}. The memory pool is updated in a momentum style with the momentum of $0.5$. For global embeddings, we sample $16384$ negative samples from the memory pool. For other hyper-parameters, we use a batch size of $224$, a learning rate of $0.03$, a temperature of $0.07$ for all three contrastive learning targets. For the pre-train, 4 NVIDIA V100 GPUs are used. The training process is divided in two steps. The first step only pre-train the model using sample-level modality-invariant representation learning target for $100$ epochs. The second stage adds the other two learning targets and trains for another $100$ epochs. The whole training process takes approximately 48 hours.

\noindent\textbf{Mixing Heterogeneous Datasets.} Since we mix several heterogeneous human datasets for pre-train, we need to mask out the missing modalities. For example, when we use NTU RGB+D and MPII for pre-train. The former dataset has all the required modalities, while the latter one misses depth maps. Therefore, for the hierarchical contrastive learning targets, we mask out the missing depth embeddings of MPII for all the positive pairs sampling. By using the masking technique, it is possible to combine multiple heterogeneous datasets into this pre-train paradigm as long as there are at least two common modalities.

\noindent\textbf{Datasets for Pre-train.} For NTU RGB+D, we only use the version with 60 actions~\cite{shahroudy2016ntu}. With the provided RGB-D videos, we uniformly sample one frame from every 30 frames, which makes $143648$ samples. The RGB and depth frames are calibrated by the correspondences provided by the 2D keypoints positions on RGB and depths. For MPII and COCO, we use the full training sets for pre-train.

\subsection{DensePose Estimation}

For the DensePose~\cite{guler2018densepose} estimation, we use the official open-sourced implementation \footnote{\url{https://github.com/facebookresearch/detectron2}}. For the full training set, we train the network for $130000$ iterations with a batch size of $16$, a learning rate of $0.01$ on 4 NVIDIA V100 GPUs, which takes around $80$ hours to train. For the $10\%$ training set, we train the network for $13000$ iterations with a learning rate of $0.005$ and other settings the same, which takes 9 hours to train. The $10\%$ training set is uniformly sampled from the default ordered training set.

\subsection{RGB Human Parsing}

For the RGB human parsing on Human3.6M~\cite{h36m_pami}, we use the official HRNet~\cite{sun2019deep} semantic segmentation implementation \footnote{\url{https://github.com/HRNet/HRNet-Semantic-Segmentation}}. Different ratios of training settings are uniformly sampled from the default ordered full training set. For the full training set, we train the network for $50$ epochs with a learning rate of $0.007$, a batch size of $40$ on 2 NVIDIA V100 GPUs. For other data-efficient settings, we train the network for $150$ epochs with other settings the same. We use the the standard dataset split protocol, where the subjects $1, 5, 6, 7, 8$ are for training and the subjects $9$ and $11$ are for evaluation.

\subsection{Depth Human Parsing}

For the depth human parsing on NTURGBD-Parsing-4K, we use the same implementation as that of RGB human parsing. To use the HRNet to encode depth maps, we repeat the depth dimension for three times to fit the RGB input, which is also how \nickname{} deals with depth inputs. For all training settings, we train the network for $150$ epochs with a learning rate of $0.007$, a batch size of $80$ on 2 NVIDIA V100 GPUs. Even though the encoder is used to deal with depth inputs, we still initialize it using ImageNet pre-train for that it might help with the performance proved by some previous works~\cite{xiong2019a2j}.

\subsection{3D Pose Estimation on Depth}

For the 3D pose estimation from depth maps on ITOP~\cite{haque2016towards}, we choose to adapt the official implementation \footnote{\url{https://github.com/zhangboshen/A2J}} of A2J~\cite{xiong2019a2j}. The original implementation uses ResNet as the backbone. And we switch to HRNet. Since the original implementation only provides validation scripts, we re-implement the whole training pipeline. We change the original normalization method where a global mean and variance is counted for a global normalization. Instead, we perform an online instance normalization where we only centralize each depth pixel to zero mean but do not normalize its variance, since its a better way to prevent the over-fitting to the relatively small dataset. We train the network for $50$ epochs with a learning rate of $0.00035$ and a batch size of $12$ on one NVIDIA V100 GPU. As for the dataset, we use the side-view of ITOP since the depth maps in pre-train are side views. Following the official dataset split, there are $17991$ samples for training and $4863$ for testing. Following the practice of A2J~\cite{xiong2019a2j}, we initialize the encoders using ImageNet pre-train.

\subsection{Cross-Modality Supervision}

To experiment with the cross-modality supervision, we choose the downstream task of human parsing on NTURGBD-Parsing-4K. The modalities to experiment with are RGB and depth. To make the experiment fair and the networks to converge faster, the backbones are initialized by CMC~\cite{tian2020contrastive} pre-train. The following descriptions are for the setting of `RGB$\rightarrow$Depth', where the source modality is RGB and the target modality is depth. To implement `Depth$\rightarrow$RGB', one can simply switch the source and target modalities. At training time, a randomly initialized segmentation header, which is the same one used for human parsing experiments, is attached to the dense mapper network of RGB. Then the network is trained with both the hierarchical contrastive learning targets $\mathcal{L}$ and a cross-entropy loss $\mathcal{L}'$ for the supervision of the segmentation. For the `No Contrastive' baseline, we only train with $\mathcal{L}'$. As for the `CMC' baseline, the network is supervised by both the learning target proposed by CMC~\cite{tian2020contrastive} $\mathcal{L}$ and the segmentation loss $\mathcal{L}'$. Note that, during the whole training time, including the CMC pre-train, the target modality of NTURGBD-Parsing-4K is not exposed to better simulate the application scenario. In order to build the connection between RGB and depth during training time, we mix the NTURGBD-Parsing-4K with NTU RGB+D which is the same one used for our pre-train. At inference time, we attach the trained segmentation head to the mapper network of depth. Since the dense embeddings of RGB and depth are aligned thanks to our hierarchical contrastive learning targets, it is reasonable for the segmentation head to be able to handle the dense embeddings of depth.

\subsection{Missing-Modality Inference}

We also use human parsing on NTURGBD-Parsing-4K to experiment with our extension of missing-modality inference. The basic setup is the same as that of the cross-modality supervision experiments. At training time, we take the dense embeddings of both RGB and depth together for a max pooling operation for a simple feature-level fusion. Then the fused dense embedding is passed to a segmentation header, which is the same one used by the human parsing experiment, to produce the segmentation prediction. The network is supervised with both the hierarchical contrastive learning targets $\mathcal{L}$ and a cross-entropy loss $\mathcal{L}'$ for segmentation supervision. Similarly, the `No contrastive' baseline does not use any contrastive learning targets. The `CMC' baseline uses the contrastive learning target proposed in CMC~\cite{tian2020contrastive} as $\mathcal{L}$. At inference time, if RGB is missing, then the dense embedding of depth is passed to the trained segmentation header for prediction. Since the dense embeddings of RGB and depth are aligned and the segmentation header is trained with the fusion of both embeddings, missing one of them will still produce reasonable predictions.

\section{More Quantitative Results} \label{sec:moreres}

\noindent\textbf{DensePose Estimation.} Due to the page limitation, we could not report all metrics for the DensePose~\cite{guler2018densepose} estimation. Therefore, we report them in this supplementary material. As shown in Tab. \ref{tab:denseposedetails}, detailed results of all settings mentioned in the main paper are listed. Specifically, for the initialization of the network, we test with the network randomly initialized (`From Scratch') and the network initialized by ImageNet pre-train (`IN Pre-train'). As for the ratio of training data, we test with the full training set and $10\%$ of the training set. As for the pre-train datasets, we test with two combinations: NTU RGB+D $+$ MPII and NTU RGB+D $+$ COCO. As for the backbone, we test with HRNet-W18 and HRNet-W32. Compared with the baseline and two other state-of-the-art pre-train counterparts, our method outperforms them in most of the metrics. Especially, our method has advantages in GPS and GPSM, which are two critical metrics for DensePose quality. Additionally, we also report full results of the ablation study. The detailed results further validates the analysis in the main paper.

\noindent\textbf{RGB Human Parsing.} We further report detailed RGB human parsing results on Human3.6M~\cite{h36m_pami} that could not fit into the main paper. As shown in Tab. \ref{tab:human36mdetails}, we report the per-class IoU for all the settings reported in the main paper. Similarly, for the initialization of the network, we test with the network randomly initialized (`From Scratch') and the network initialized by ImageNet pre-train (`IN Pre-train'). As for the ratio of training data, we test with the full training set, $20\%$, $10\%$ and $1\%$ of the training set. The pre-train datasets are NTU RGB+D $+$ MPII. In most classes, our method outperforms comparison methods. Moreover, we also report per-class IoU for the four settings in ablation study, which are in line with our analysis in the main paper.

\noindent\textbf{Depth Human Parsing.} We report detailed depth human parsing results on NTURGBD-Parsing-4K. As shown in Tab. \ref{tab:nturgbddetails}, we report the per-class IoU for all the settings reported in the main paper. We initialize the networks using ImageNet pre-train. Two ratios of the training set, \ie full and $20\%$, are tested. We also change the backbone to PointNet++~\cite{qi2017pointnet++} (`PN++'). Since it is a point-based backbone, the `background' class is ignored and not included in the calculation of mIoU. The per-class IoU results also agree with the conclusion in the main paper that our method is superior than other comparison methods.

\noindent\textbf{Cross-Modality Supervision.} As shown in Tab. \ref{tab:appdetails}, we report detailed per-class IoU for the experiments of cross-modality supervision. In both `RGB$\rightarrow$Depth' and `Depth$\rightarrow$RGB' settings, our method outperforms other baseline methods in all classes. Especially, other baseline methods barely make correct predictions while ours makes a huge improvement.

\noindent\textbf{Missing-Modality Inference.} As shown in Tab. \ref{tab:appdetails}, we list detailed per-class IoU for the experiments of missing-modality inference. In both `Only RGB' and `Only Depth' settings, our method outperforms baseline methods in most classes. Therefore, the detailed results further validates the conclusions made in the main paper.

\section{More Qualitative Results} \label{sec:vis}

\begin{figure*}[t]
    \begin{center}
        \includegraphics[width=0.9\linewidth]{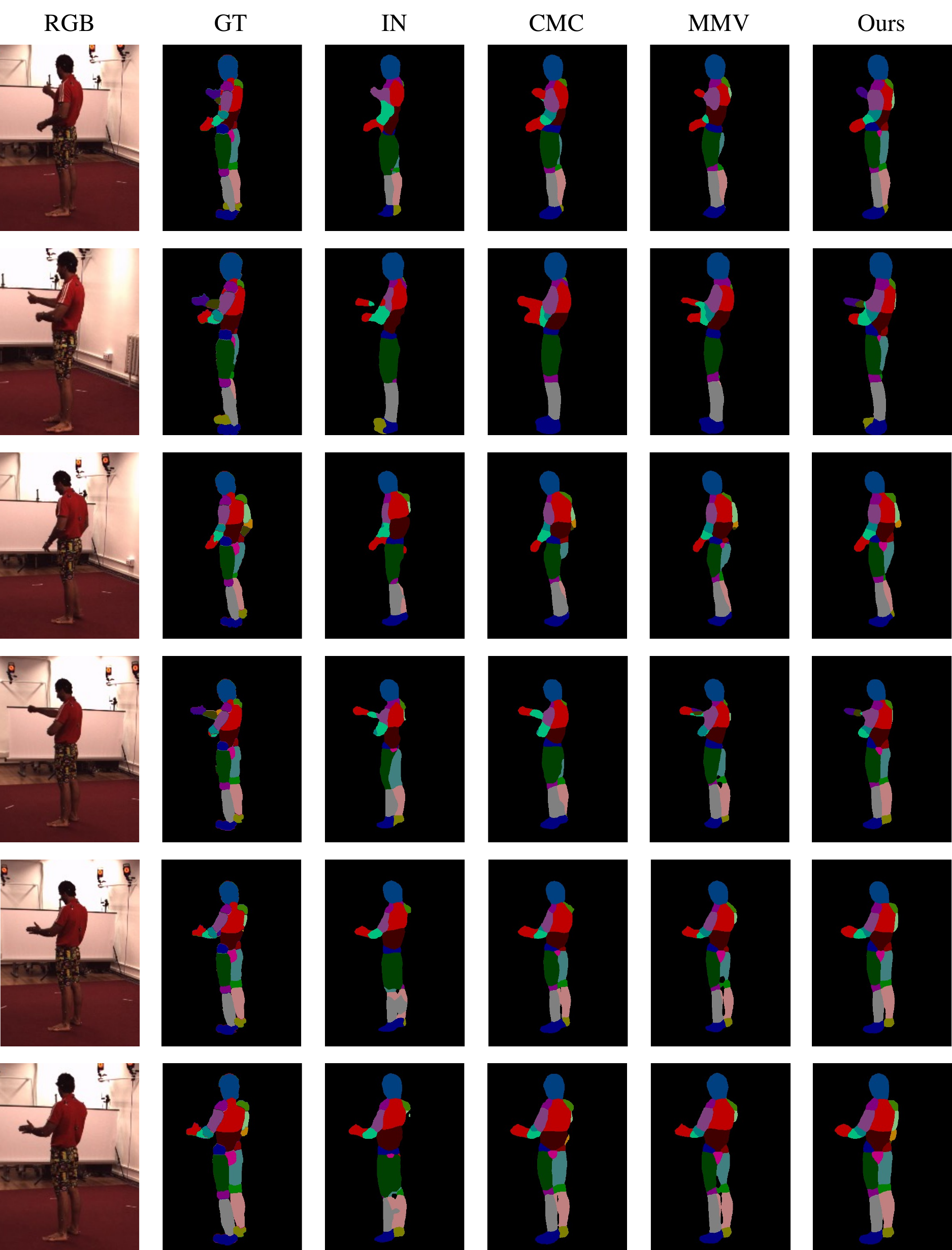}
    \end{center}
    \caption{Qualitative Results of RGB Human Parsing on Human3.6M with $10\%$ of the Training Set.}
    \label{fig:10_human36m}
\end{figure*}

\begin{figure*}[t]
    \begin{center}
        \includegraphics[width=0.9\linewidth]{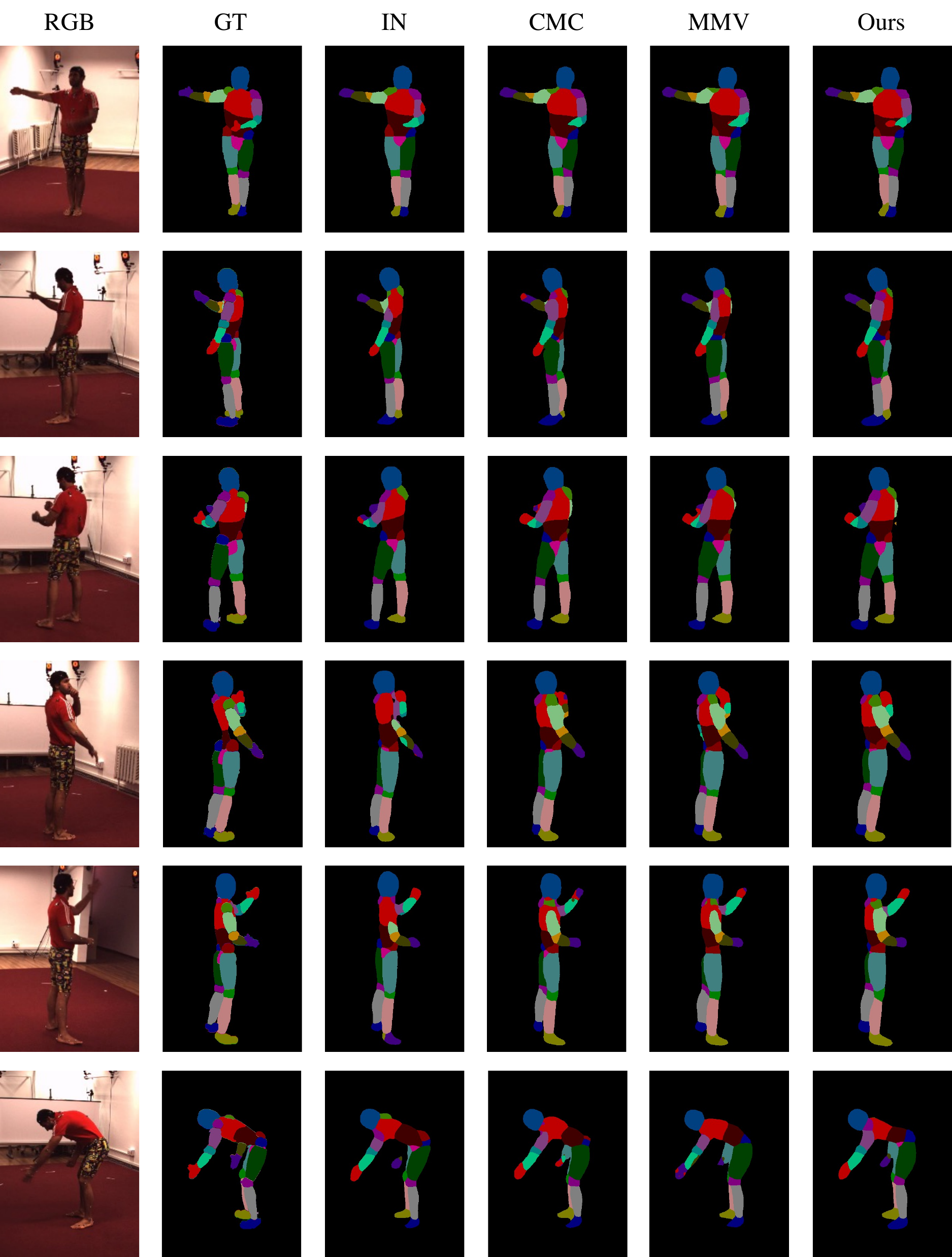}
    \end{center}
    \caption{Qualitative Results of RGB Human Parsing on Human3.6M with the Full Training Set.}
    \label{fig:100_human36m}
\end{figure*}

\begin{figure*}[t]
    \begin{center}
        \includegraphics[width=0.9\linewidth]{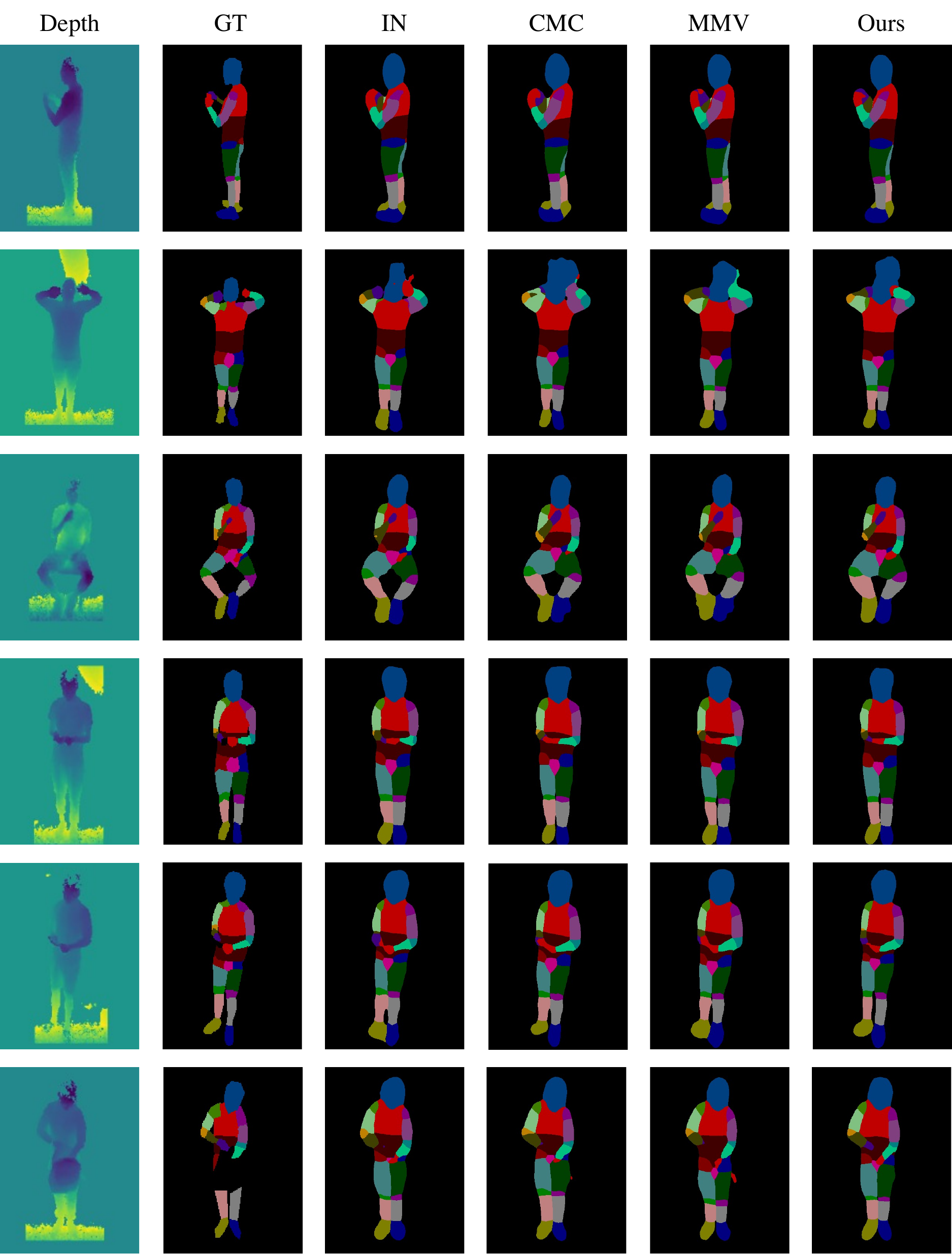}
    \end{center}
    \caption{Qualitative Results of Depth Human Parsing on NTURGBD-Parsing-4K with the Full Training Set.}
    \label{fig:v2_nturgbd}
\end{figure*}

More qualitative results of RGB human parsing on Human3.6M~\cite{h36m_pami} and depth human parsing on NTURGBD-Parsing-4K are shown in Fig. \ref{fig:10_human36m}, Fig \ref{fig:100_human36m} and Fig. \ref{fig:v2_nturgbd}. We choose to visualize both the full training set and $10\%$ training set for RGB human parsing. The segmentation results produced by our pre-train model are superior than those of other comparison methods, especially in data-efficient settings. For challenging classes like hands and elbows, our method is capable of producing correct predictions constantly while other methods struggle. The depth map is a challenging modality for the dense prediction task like semantic segmentation. Our method manages to produce reasonable predictions that are better than those of other comparison methods.


\clearpage
\begin{sidewaystable}[!htbp]
    \vspace{250pt}
    \caption{Detailed DensePose Estimation Results on COCO. `Ratio' stands for the ratio of training data for downstream tasks transfer. \textsuperscript{*} randomly initializes the model before pre-training. $^\dagger$ initializes the model by ImageNet pre-train before pre-training. `Ablation1' is `Sample-level Mod-invariant'; `Abation2' is `+ Hard Dense Intra-sample'; `Ablation3' is `+ Soft Dense Intra-sample'; `Ablation4' is `+ Sparse Structure-aware'. $^\ddagger$ uses HRNet-W32 while others all use HRNet-W18. All results in [\%].}
    \vspace{-10pt}
    \begin{center}
    \tiny{
        \addtolength{\tabcolsep}{-3.9pt}
        \begin{tabular}{l|c|c|cccccc|cccccccccc|cccccccccc|cccccccccc}
            \Xhline{1pt}
                \multirow{2}{*}{Methods} & \multirow{2}{*}{Pre-train Datasets} & \multirow{2}{*}{Ratio} & \multicolumn{6}{c|}{BBox} & \multicolumn{10}{c|}{GPS} & \multicolumn{10}{c|}{GPSM} & \multicolumn{10}{c}{IoU} \\
                & & & AP & AP50 & AP75 & APs & APm & APl & AP & AP50 & AP75 & APm & APl & AR & AR50 & AR75 & ARm & ARl & AP & AP50 & AP75 & APm & APl & AR & AR50 & AR75 & ARm & ARl & AP & AP50 & AP75 & APm & APl & AR & AR50 & AR75 & ARm & ARl \\
                \hline \hline
                From Scratch & - & 100\%                                                 & 57.27 & 85.44 & 61.37 & 28.31 & 53.64 & 72.46 & 62.03 & 89.93 & 69.57 & 58.29 & 62.89 & 69.00 & 93.05 & 76.06 & 60.78 & 69.55 & 63.61 & 91.89 & 74.56 & 57.18 & 64.71 & 69.24 & 95.05 & 79.36 & 59.72 & 69.88 & 65.88 & 93.86 & 77.95 & 58.43 & 67.04 & 71.11 & 96.12 & 82.97 & 60.35 & 71.83 \\
                CMC\textsuperscript{*}~\cite{tian2020contrastive} & NTURGBD+MPII & 100\% & 57.27 & 85.44 & 61.37 & 28.31 & 53.64 & 72.46 & 62.03 & 89.93 & 69.57 & 58.29 & 62.89 & 69.00 & 93.05 & 76.06 & 60.78 & 69.55 & 63.61 & 91.89 & 74.56 & 57.18 & 64.71 & 69.24 & 95.05 & 79.36 & 59.72 & 69.88 & 65.88 & 93.86 & 77.95 & 58.43 & 67.04 & 71.11 & 96.12 & 82.97 & 60.35 & 71.83 \\
                MMV\textsuperscript{*}~\cite{alayrac2020self} & NTURGBD+MPII & 100\%     & 60.33 & 86.82 & 66.23 & 30.37 & 57.29 & 75.59 & 64.97 & 91.10 & 72.99 & 59.59 & 65.90 & 71.57 & 93.80 & 79.00 & 61.42 & 72.25 & 65.66 & 92.22 & 77.31 & 57.81 & 66.81 & 70.80 & 94.96 & 81.77 & 59.72 & 71.54 & 66.96 & 93.80 & 79.14 & 59.73 & 67.97 & 71.83 & 95.72 & 83.68 & 61.42 & 72.52 \\
                Ours\textsuperscript{*} & NTURGBD+MPII & 100\%                           & \textbf{61.33} & \textbf{87.81} & \textbf{66.48} & \textbf{31.80} & \textbf{58.27} & \textbf{76.30} & \textbf{65.89} & \textbf{92.20} & \textbf{75.36} & \textbf{61.12} & \textbf{66.87} & \textbf{72.52} & \textbf{94.69} & \textbf{81.01} & \textbf{62.91} & \textbf{73.17} & \textbf{66.92} & \textbf{93.27} & \textbf{78.72} & \textbf{59.75} & \textbf{67.97} & \textbf{71.87} & \textbf{95.50} & \textbf{82.92} & \textbf{61.63} & \textbf{72.56} & \textbf{67.66} & \textbf{94.41} & \textbf{80.14} & \textbf{61.08} & \textbf{68.71} & \textbf{72.62} & \textbf{96.21} & \textbf{84.80} & \textbf{62.62} & \textbf{73.29} \\
                \hline \hline
                IN Pre-train & - & 100\%                                        & 62.66 & \textbf{89.28} & 68.47 & \textbf{35.78} & 59.59 & 76.47 & 66.49 & 92.11 & 75.47 & \textbf{64.45} & 67.43 & 73.41 & 94.87 & 81.28 & \textbf{66.24} & 73.89 & 67.42 & 93.20 & 79.72 & \textbf{62.11} & 68.52 & 72.63 & 95.90 & 83.82 & \textbf{63.97} & 73.20 & 68.63 & 94.64 & 81.81 & \textbf{62.56} & 69.72 & 73.45 & 96.30 & 85.64 & \textbf{64.11} & 74.08 \\
                CMC$^\dagger$~\cite{tian2020contrastive} & NTURGBD+MPII & 100\% & 62.76 & 88.32 & 68.32 & 33.67 & 60.22 & 76.95 & 66.17 & 92.45 & 74.79 & 62.79 & 66.96 & 72.75 & 94.69 & 80.34 & 64.68 & 73.29 & 67.31 & 93.51 & 79.71 & 61.55 & 68.27 & 72.23 & 95.68 & 83.73 & 63.55 & 72.80 & 68.07 & 94.68 & 81.41 & 61.89 & 69.01 & 72.90 & 96.30 & 85.51 & 63.62 & 73.53 \\
                MMV$^\dagger$~\cite{alayrac2020self} & NTURGBD+MPII & 100\%     & 62.97 & 88.75 & 68.91 & 34.62 & \textbf{60.83} & 76.87 & 66.67 & 92.42 & 76.24 & 63.09 & 67.64 & 73.03 & 94.78 & 81.28 & 64.04 & 73.63 & 67.51 & 93.44 & 79.72 & 60.91 & 68.50 & 72.29 & 95.81 & 83.10 & 61.99 & 72.97 & 68.29 & 94.52 & \textbf{81.84} & 61.04 & 69.20 & 72.95 & \textbf{96.34} & 85.42 & 62.20 & 73.67 \\
                Ours$^\dagger$ & NTURGBD+MPII & 100\%                           & \textbf{63.11} & 88.66 & \textbf{69.64} & 34.53 & 60.80 & \textbf{77.01} & \textbf{67.33} & \textbf{93.20} & \textbf{76.27} & 62.63 & \textbf{68.20} & \textbf{73.77} & \textbf{95.23} & \textbf{81.59} & 63.83 & \textbf{74.44} & \textbf{68.12} & \textbf{94.09} & \textbf{79.90} & 61.35 & \textbf{68.99} & \textbf{72.94} & \textbf{96.08} & \textbf{83.95} & 62.48 & \textbf{73.64} & \textbf{68.72} & \textbf{94.74} & 81.67 & 61.50 & \textbf{69.77} & \textbf{73.47} & 96.26 & \textbf{85.73} & 62.70 & \textbf{74.19} \\
                \hline \hline
                CMC$^\dagger$~\cite{tian2020contrastive} & NTURGBD+COCO & 100\% & \textbf{63.58} & \textbf{88.94} & \textbf{69.69} & \textbf{35.24} & \textbf{61.37} & \textbf{77.46} & 67.22 & 92.68 & 76.27 & \textbf{64.61} & 68.20 & 73.75 & 95.10 & 81.59 & \textbf{65.60} & 74.29 & 67.77 & \textbf{93.65} & 79.56 & 62.62 & 68.81 & 72.78 & \textbf{95.99} & 83.55 & 63.83 & 73.38 & 68.46 & 93.93 & \textbf{81.80} & \textbf{62.60} & 69.50 & \textbf{73.37} & 95.94 & \textbf{85.78} & \textbf{63.76} & \textbf{74.02} \\
                Ours$^\dagger$ & NTURGBD+COCO & 100\%                           & 62.95 & 88.78 & 68.83 & 34.77 & 60.59 & 77.02 & \textbf{67.77} & \textbf{93.18} & \textbf{77.13} & 64.02 & \textbf{68.63} & \textbf{74.15} & \textbf{95.23} & \textbf{82.39} & 65.18 & \textbf{74.75} & \textbf{68.29} & 93.60 & \textbf{80.53} & \textbf{62.77} & \textbf{69.23} & \textbf{73.21} & \textbf{95.99} & \textbf{84.44} & \textbf{64.11} & \textbf{73.81} & \textbf{68.63} & \textbf{94.53} & 81.53 & 62.36 & \textbf{69.51} & 73.30 & \textbf{96.17} & 85.69 & 63.33 & 73.97 \\
                \hline \hline
                From Scratch & - & 10\%                                                 & 39.38 & 72.29 & 37.98 & 12.03 & 35.59 & 55.16 & 35.75 & 73.78 & 30.07 & 27.19 & 37.28 & 45.32 & 81.41 & 43.69 & 31.49 & 46.25 & 41.62 & 80.25 & 38.81 & 30.71 & 43.33 & 49.67 & 87.34 & 50.74 & 35.60 & 50.61 & 49.92 & 85.96 & 54.21 & 38.56 & 51.61 & 57.90 & 91.62 & 65.14 & 43.19 & 58.88 \\
                CMC\textsuperscript{*}~\cite{tian2020contrastive} & NTURGBD+MPII & 10\% & 44.92 & 76.41 & 46.34 & 16.09 & 41.58 & 60.88 & 43.84 & 79.94 & 42.97 & 40.31 & 45.00 & 52.25 & 85.69 & 54.21 & 42.55 & 52.90 & 47.94 & 84.22 & 50.23 & 41.42 & 49.33 & 55.00 & 89.43 & 59.61 & 43.90 & 55.74 & 54.00 & 88.65 & 61.20 & 46.16 & 55.45 & 60.96 & 92.73 & 69.82 & 48.94 & 61.77 \\
                MMV\textsuperscript{*}~\cite{alayrac2020self} & NTURGBD+MPII & 10\%     & 43.24 & 74.53 & 44.27 & 13.49 & 40.43 & 59.22 & 41.40 & 77.54 & 39.26 & 34.54 & 42.93 & 50.45 & 84.22 & 51.67 & 37.02 & 51.35 & 45.99 & 81.83 & 48.11 & 37.22 & 47.64 & 53.58 & 88.19 & 58.18 & 39.72 & 54.51 & 52.52 & 87.73 & 59.10 & 43.45 & 54.01 & 59.80 & 92.15 & 68.26 & 46.17 & 60.71 \\
                Ours\textsuperscript{*} & NTURGBD+MPII & 10\%                           & \textbf{47.76} & \textbf{78.66} & \textbf{50.56} & \textbf{18.59} & \textbf{45.19} & \textbf{63.12} & \textbf{48.47} & \textbf{82.74} & \textbf{51.27} & \textbf{45.21} & \textbf{49.51} & \textbf{56.41} & \textbf{87.56} & \textbf{61.21} & \textbf{47.73} & \textbf{56.99} & \textbf{51.65} & \textbf{85.50} & \textbf{56.06} & \textbf{45.86} & \textbf{52.76} & \textbf{58.27} & \textbf{90.33} & \textbf{64.20} & \textbf{48.44} & \textbf{58.91} & \textbf{56.15} & \textbf{89.63} & \textbf{64.34} & \textbf{49.27} & \textbf{57.43} & \textbf{62.92} & \textbf{93.36} & \textbf{72.54} & \textbf{51.13} & \textbf{63.71} \\
                \hline \hline
                IN Pre-train & - & 10\%                                        & 48.29 & 80.07 & 50.90 & 19.55 & 44.54 & 63.64 & 44.34 & 78.77 & 44.83 & 36.85 & 46.49 & 54.54 & 86.49 & 57.38 & 39.29 & 55.56 & 49.11 & 84.02 & 51.91 & 40.57 & 51.38 & 57.78 & 90.82 & 63.17 & 43.40 & 58.73 & 56.11 & 88.10 & 63.14 & 46.91 & 58.16 & 63.73 & 92.56 & 72.94 & 49.50 & 64.69 \\
                CMC$^\dagger$~\cite{tian2020contrastive} & NTURGBD+MPII & 10\% & 49.21 & 80.43 & 52.01 & 21.66 & 47.22 & 63.71 & 48.82 & 83.85 & 51.39 & 43.97 & 50.05 & 57.21 & 88.72 & 61.35 & 46.67 & 57.92 & 52.57 & 86.67 & 57.21 & 45.69 & 53.89 & 59.52 & 91.26 & 66.21 & 48.65 & 60.25 & 57.94 & \textbf{90.82} & 66.11 & 50.11 & 59.30 & 64.80 & \textbf{94.07} & 74.50 & 52.34 & 65.64 \\
                MMV$^\dagger$~\cite{alayrac2020self} & NTURGBD+MPII & 10\%     & 50.16 & \textbf{81.33} & \textbf{53.68} & 22.17 & 47.73 & 64.76 & 50.28 & 84.02 & 53.95 & 44.06 & 51.68 & 58.43 & 89.03 & 63.40 & 45.74 & 59.28 & 53.54 & 86.66 & 59.25 & 44.91 & 55.06 & 60.35 & 91.22 & 67.59 & 46.74 & 61.25 & 58.32 & 90.15 & 67.10 & 49.21 & 59.89 & \textbf{64.94} & 93.62 & 74.90 & 50.92 & \textbf{65.88} \\
                Ours$^\dagger$ & NTURGBD+MPII & 10\%                           & \textbf{50.29} & 80.94 & 53.62 & \textbf{22.74} & \textbf{47.91} & \textbf{65.25} & \textbf{51.50} & \textbf{84.89} & \textbf{54.64} & \textbf{45.63} & \textbf{52.68} & \textbf{59.03} & \textbf{89.17} & \textbf{63.58} & \textbf{47.80} & \textbf{59.78} & \textbf{54.47} & \textbf{87.13} & \textbf{60.26} & \textbf{46.82} & \textbf{55.71} & \textbf{60.66} & \textbf{91.44} & \textbf{67.99} & \textbf{49.29} & \textbf{61.42} & \textbf{58.66} & 90.32 & \textbf{68.82} & \textbf{50.63} & \textbf{59.93} & \textbf{64.94} & 93.89 & \textbf{75.57} & \textbf{52.48} & 65.77 \\
                \hline \hline
                CMC$^\dagger$~\cite{tian2020contrastive} & NTURGBD+COCO & 10\% & 51.77 & 81.86 & 55.93 & 23.64 & 49.54 & 66.50 & 53.53 & 86.23 & 58.30 & 46.79 & 54.78 & 60.81 & 90.15 & 66.47 & 50.21 & 61.52 & 56.18 & 88.58 & 64.41 & 48.38 & 57.39 & 62.35 & 92.33 & 71.73 & \textbf{52.13} & 63.04 & 59.37 & 90.74 & 68.92 & \textbf{52.20} & 60.62 & 65.83 & 93.85 & 76.42 & \textbf{54.68} & 66.58 \\
                Ours$^\dagger$ & NTURGBD+COCO & 10\%                           & \textbf{52.18} & \textbf{82.71} & \textbf{56.57} & \textbf{24.92} & \textbf{49.96} & \textbf{66.86} & \textbf{54.01} & \textbf{86.41} & \textbf{58.31} & \textbf{48.15} & \textbf{55.14} & \textbf{61.53} & \textbf{90.55} & \textbf{67.10} & \textbf{50.99} & \textbf{62.24} & \textbf{56.64} & \textbf{88.67} & \textbf{64.79} & \textbf{48.78} & \textbf{57.88} & \textbf{63.05} & \textbf{92.69} & \textbf{72.18} & 51.70 & \textbf{63.81} & \textbf{59.93} & \textbf{91.31} & \textbf{69.99} & 52.01 & \textbf{61.08} & \textbf{66.50} & \textbf{94.34} & \textbf{77.40} & 54.54 & \textbf{67.29} \\
                \hline \hline
                Ablation1 & NTURGBD+MPII & 10\% & 49.21 & 80.43 & 52.01 & 21.66 & 47.22 & 63.71 & 48.82 & 83.85 & 51.39 & 43.97 & 50.05 & 57.21 & 88.72 & 61.35 & 46.67 & 57.92 & 52.57 & 86.67 & 57.21 & 45.69 & 53.89 & 59.52 & 91.26 & 66.21 & 48.65 & 60.25 & 57.94 & \textbf{90.82} & 66.11 & 50.11 & 59.30 & 64.80 & \textbf{94.07} & 74.50 & 52.34 & 65.64 \\
                Ablation2 & NTURGBD+MPII & 10\% & 49.40 & 80.93 & 51.61 & 21.94 & 47.56 & 63.72 & 49.14 & 82.17 & 51.83 & 44.53 & 50.43 & 57.65 & 87.83 & 62.42 & 46.88 & 58.37 & 52.49 & 85.69 & 58.42 & 46.56 & 53.84 & 59.61 & 90.68 & 66.87 & 49.15 & 60.31 & 57.30 & 89.93 & 65.38 & 50.58 & 58.74 & 64.29 & 93.80 & 73.96 & \textbf{53.12} & 65.03 \\
                Ablation3 & NTURGBD+MPII & 10\% & 50.21 & \textbf{81.02} & 53.39 & \textbf{22.86} & 47.78 & 64.75 & 50.25 & 84.20 & 53.58 & 44.74 & 51.41 & 57.95 & 88.50 & 62.91 & 47.66 & 58.64 & 53.42 & 86.41 & 58.95 & 46.27 & 54.68 & 59.85 & 90.68 & 66.87 & 49.01 & 60.56 & 57.70 & 89.75 & 67.40 & \textbf{50.95} & 58.89 & 64.40 & 93.31 & 74.72 & 52.91 & 65.17 \\
                Ablation4 & NTURGBD+MPII & 10\% & \textbf{50.29} & 80.94 & \textbf{53.62} & 22.74 & \textbf{47.91} & \textbf{65.25} & \textbf{51.50} & \textbf{84.89} & \textbf{54.64} & \textbf{45.63} & \textbf{52.68} & \textbf{59.03} & \textbf{89.17} & \textbf{63.58} & \textbf{47.80} & \textbf{59.78} & \textbf{54.47} & \textbf{87.13} & \textbf{60.26} & \textbf{46.82} & \textbf{55.71} & \textbf{60.66} & \textbf{91.44} & \textbf{67.99} & \textbf{49.29} & \textbf{61.42} & \textbf{58.66} & 90.32 & \textbf{68.82} & 50.63 & \textbf{59.93} & \textbf{64.94} & 93.89 & \textbf{75.57} & 52.48 & \textbf{65.77} \\
                \hline \hline
                IN Pre-train$^\ddagger$ & - & 10\%                              & \textbf{55.10} & \textbf{84.95} & \textbf{59.81} & \textbf{27.33} & 52.01 & \textbf{69.65} & 54.60 & 86.42 & 59.80 & 48.63 & 55.93 & 62.48 & 90.68 & 68.12 & 50.43 & 63.29 & 57.60 & 88.62 & 66.39 & 49.61 & 59.15 & 64.09 & 92.55 & 72.49 & 51.63 & \textbf{64.92} & 61.73 & \textbf{91.74} & 72.52 & 53.24 & \textbf{63.10} & \textbf{67.99} & \textbf{94.69} & 78.73 & 55.11 & \textbf{68.86} \\
                CMC$^\ddagger$~\cite{tian2020contrastive} & NTURGBD+MPII & 10\% & 53.88 & 83.61 & 58.47 & 26.16 & 51.24 & 68.64 & 54.62 & 86.83 & 58.93 & 47.45 & 55.86 & 62.05 & 90.73 & 67.19 & 49.01 & 62.93 & 57.46 & 88.57 & 65.55 & 49.59 & 58.71 & 63.57 & 92.33 & 72.05 & 51.21 & 64.39 & 61.14 & 91.65 & 72.58 & 53.46 & 62.39 & 67.21 & 94.25 & 78.73 & 54.96 & 68.03 \\
                Ours$^\ddagger$ & NTURGBD+MPII & 10\%                           & 54.55 & 83.77 & 58.83 & 26.94 & \textbf{52.56} & 68.78 & \textbf{55.80} & \textbf{87.37} & \textbf{61.22} & \textbf{51.83} & \textbf{56.75} & \textbf{62.70} & \textbf{90.91} & \textbf{68.48} & \textbf{53.12} & \textbf{63.34} & \textbf{58.36} & \textbf{89.29} & \textbf{67.31} & \textbf{52.03} & \textbf{59.39} & \textbf{64.11} & \textbf{92.87} & \textbf{73.52} & \textbf{53.55} & 64.82 & \textbf{61.75} & 91.59 & \textbf{72.79} & \textbf{54.77} & 62.87 & 67.72 & 94.38 & \textbf{79.00} & \textbf{56.38} & 68.48 \\
            \Xhline{1pt}
        \end{tabular}
    }
    \end{center}
    \label{tab:denseposedetails}
\end{sidewaystable}

\clearpage
\begin{sidewaystable}[!htbp]
    \vspace{250pt}
    \caption{Detailed Human Parsing Results on Human3.6M. `Ratio' stands for the ratio of training data for downstream tasks transfer. \textsuperscript{*} randomly initializes the model before pre-training. $^\dagger$ initializes the model by ImageNet pre-train before pre-training. `Ablation1' is `Sample-level Mod-invariant'; `Abation2' is `+ Hard Dense Intra-sample'; `Ablation3' is `+ Soft Dense Intra-sample'; `Ablation4' is `+ Sparse Structure-aware'. All results in [\%].}
    \vspace{-10pt}
    \begin{center}
    \tiny{
        \addtolength{\tabcolsep}{-3.9pt}
        \begin{tabular}{l|c|ccccccccccccccccccccccccc|c}
            \Xhline{1pt}
            Methods & Ratio & bg & right hip & right knee & right foot & left hip & left knee & left foot & left shoulder & left elbow & left hand & right shoulder & right elbow & right hand & crotch & right thigh & right calf & left thigh & left calf & lower spine & upper spine & head & left arm & left forearm & right arm & right forearm & mIoU \\
            \hline \hline
            From Scratch & 100\%                                      & 99.53 & 29.66 & 32.28 & 52.94 & 25.86 & 30.71 & 49.59 & 35.14 & 20.78 & 29.46 & 34.33 & 19.01 & 30.61 & 37.24 & 54.53 & 65.77 & 54.35 & 64.12 & 50.56 & 57.39 & 77.57 & 40.71 & 35.83 & 38.14 & 36.99 & 44.13 \\
            CMC\textsuperscript{*}~\cite{tian2020contrastive} & 100\% & 99.55 & 44.95 & 48.96 & 54.05 & 37.57 & 41.55 & 53.41 & 47.98 & 33.35 & 37.05 & 46.01 & 32.28 & 38.18 & 50.81 & 64.84 & 69.95 & 62.22 & 68.16 & 67.76 & 71.78 & 81.23 & 56.32 & 46.75 & 55.92 & 47.53 & 54.33 \\
            MMV\textsuperscript{*}~\cite{alayrac2020self} & 100\%     & 99.54 & 39.92 & 45.88 & 53.18 & 34.04 & 34.77 & 52.97 & 45.35 & 31.22 & 34.10 & 45.08 & 31.92 & 35.26 & 53.11 & 63.96 & 69.41 & 60.81 & 66.78 & 67.53 & 72.35 & 79.96 & 55.63 & 43.48 & 56.07 & 44.99 & 52.69 \\
            Ours\textsuperscript{*} & 100\%                           & \textbf{99.61} & \textbf{51.09} & \textbf{58.38} & \textbf{61.22} & \textbf{40.81} & \textbf{48.02} & \textbf{59.65} & \textbf{54.21} & \textbf{44.75} & \textbf{48.57} & \textbf{52.78} & \textbf{43.66} & \textbf{48.68} & \textbf{56.70} & \textbf{71.43} & \textbf{75.89} & \textbf{67.98} & \textbf{73.73} & \textbf{71.63} & \textbf{76.79} & \textbf{83.89} & \textbf{64.64} & \textbf{57.51} & \textbf{64.47} & \textbf{57.97} & \textbf{61.36} \\
            \hline \hline
            From Scratch & 20\%                                      & 99.52 & 29.07 & 32.15 & 50.17 & 23.11 & 24.90 & 48.12 & 34.98 & 17.26 & 23.58 & 33.20 & 17.75 & 25.39 & 34.76 & 54.76 & 65.22 & 51.44 & 62.03 & 55.16 & 60.63 & 78.27 & 39.47 & 30.30 & 37.31 & 31.84 & 42.41 \\
            CMC\textsuperscript{*}~\cite{tian2020contrastive} & 20\% & 99.53 & 41.21 & 42.49 & 49.99 & 39.44 & 39.37 & 49.24 & 46.58 & 30.02 & 33.19 & 45.35 & 30.08 & 34.48 & 53.06 & 62.33 & 66.06 & 61.24 & 64.88 & 67.02 & 71.59 & 80.35 & 54.80 & 42.67 & 54.03 & 43.61 & 52.10 \\
            MMV\textsuperscript{*}~\cite{alayrac2020self} & 20\%     & 99.51 & 38.19 & 43.71 & 49.45 & 34.44 & 32.47 & 49.59 & 42.88 & 28.15 & 30.74 & 43.23 & 28.81 & 32.13 & 52.26 & 62.83 & 66.82 & 59.55 & 63.94 & 66.44 & 71.66 & 79.42 & 54.10 & 40.48 & 54.03 & 41.73 & 50.66 \\
            Ours\textsuperscript{*} & 20\%                           & \textbf{99.59} & \textbf{46.76} & \textbf{51.17} & \textbf{56.35} & \textbf{43.52} & \textbf{48.66} & \textbf{55.20} & \textbf{53.79} & \textbf{42.58} & \textbf{43.76} & \textbf{52.41} & \textbf{42.29} & \textbf{44.58} & \textbf{60.19} & \textbf{68.57} & \textbf{71.95} & \textbf{67.80} & \textbf{70.67} & \textbf{70.56} & \textbf{75.74} & \textbf{82.68} & \textbf{62.90} & \textbf{52.71} & \textbf{62.44} & \textbf{52.48} & \textbf{59.17} \\
            \hline \hline
            From Scratch & 10\%                                      & 99.39 & 18.84 & 19.95 & 35.96 & 19.55 & 15.23 & 32.38 & 23.23 & 12.43 & 15.57 & 20.73 & 12.46 & 18.64 & 28.46 & 40.41 & 45.51 & 37.85 & 42.10 & 51.25 & 58.41 & 74.93 & 25.77 & 20.47 & 23.29 & 22.50 & 32.61 \\
            CMC\textsuperscript{*}~\cite{tian2020contrastive} & 10\% & 99.49 & 35.71 & 37.05 & 44.08 & 36.65 & 33.37 & 42.36 & 44.43 & 23.56 & 29.01 & 43.30 & 24.51 & 30.05 & 52.29 & 60.05 & 60.32 & 58.81 & 58.03 & 65.01 & 70.68 & 79.19 & 52.11 & 37.89 & 51.56 & 39.64 & 48.37 \\
            MMV\textsuperscript{*}~\cite{alayrac2020self} & 10\%     & 99.46 & 33.84 & 34.77 & 42.58 & 33.47 & 27.94 & 41.83 & 37.89 & 21.06 & 26.17 & 38.65 & 23.36 & 27.40 & 50.05 & 58.14 & 59.71 & 56.35 & 57.01 & 64.47 & 69.99 & 77.38 & 50.43 & 35.00 & 51.34 & 37.35 & 46.23 \\
            Ours\textsuperscript{*} & 10\%                           & \textbf{99.56} & \textbf{42.65} & \textbf{48.60} & \textbf{52.82} & \textbf{44.08} & \textbf{45.44} & \textbf{51.58} & \textbf{52.21} & \textbf{40.08} & \textbf{41.70} & \textbf{51.70} & \textbf{40.35} & \textbf{42.73} & \textbf{59.06} & \textbf{66.85} & \textbf{69.09} & \textbf{66.78} & \textbf{67.47} & \textbf{68.73} & \textbf{74.82} & \textbf{81.39} & \textbf{61.15} & \textbf{49.08} & \textbf{60.39} & \textbf{48.64} & \textbf{57.08} \\
            \hline \hline
            From Scratch & 1\%                                      & 98.39 & 0.00 & 0.00 & 0.00 & 0.00 & 0.00 & 0.00 & 0.00 & 0.00 & 0.00 & 0.00 & 0.00 & 0.00 & 0.00 & 18.12 & 0.00 & 15.94 & 0.00 & 18.71 & 30.70 & 0.00 & 0.00 & 0.00 & 0.00 & 0.00 & 7.27 \\
            CMC\textsuperscript{*}~\cite{tian2020contrastive} & 1\% & 98.95 & 0.00 & 0.00 & 4.08 & 0.00 & 0.00 & 2.69 & 0.00 & 0.00 & 0.00 & 0.00 & 0.00 & 0.00 & 7.38 & 22.13 & 20.40 & 22.19 & 17.72 & 48.99 & 52.65 & 68.09 & 0.00 & 0.00 & 0.00 & 0.00 & 14.61 \\
            MMV\textsuperscript{*}~\cite{alayrac2020self} & 1\%     & 98.62 & 0.00 & 0.00 & 0.00 & 0.00 & 0.00 & 0.00 & 0.00 & 0.00 & 0.00 & 0.00 & 0.00 & 0.00 & 0.00 & 22.03 & 12.54 & 18.33 & 9.34 & 46.48 & 54.31 & 59.75 & 0.00 & 0.00 & 0.00 & 0.00 & 12.86 \\
            Ours\textsuperscript{*} & 1\%                           & \textbf{99.02} & 0.00 & 0.00 & \textbf{9.95} & 0.00 & 0.00 & \textbf{8.70} & 0.00 & 0.00 & 0.00 & 0.00 & 0.00 & 0.00 & \textbf{9.97} & \textbf{26.15} & \textbf{25.71} & \textbf{24.97} & \textbf{24.00} & \textbf{57.31} & \textbf{58.61} & \textbf{69.24} & 0.00 & 0.00 & 0.00 & 0.00 & \textbf{16.55} \\
            \hline \hline
            IN Pre-train & 100\%                             & 99.60 & 37.98 & 52.94 & 56.49 & 32.24 & 46.97 & 56.29 & 46.05 & 47.55 & 42.50 & 42.81 & 47.05 & 42.88 & 46.57 & 66.69 & \textbf{75.49} & 64.51 & \textbf{74.23} & 61.85 & 67.29 & 82.69 & 58.41 & 57.98 & 56.56 & \textbf{58.90} & 56.90 \\
            CMC$^\dagger$~\cite{tian2020contrastive} & 100\% & 99.60 & 44.36 & 51.60 & 57.77 & 41.26 & \textbf{51.46} & 54.51 & 53.20 & 42.49 & 42.28 & 51.82 & 41.57 & 41.95 & 55.85 & 68.72 & 72.62 & 67.34 & 72.18 & 69.21 & 74.36 & 82.78 & 63.95 & 54.88 & 63.38 & 54.21 & 58.93 \\
            MMV$^\dagger$~\cite{alayrac2020self} & 100\%     & 99.60 & 44.58 & 54.97 & 59.49 & 40.01 & 48.30 & 56.47 & 51.56 & 43.01 & 43.60 & 50.11 & 42.12 & 42.97 & 56.61 & 69.42 & 74.69 & 67.35 & 72.90 & 68.54 & 73.46 & 82.52 & 63.02 & 55.04 & 62.54 & 54.13 & 59.08 \\
            Ours$^\dagger$ & 100\%                           & \textbf{99.62} & \textbf{50.34} & \textbf{57.06} & \textbf{60.78} & \textbf{42.96} & 50.52 & \textbf{57.75} & \textbf{56.66} & \textbf{48.56} & \textbf{48.11} & \textbf{56.50} & \textbf{47.64} & \textbf{47.86} & \textbf{60.06} & \textbf{72.46} & 75.23 & \textbf{70.20} & 73.58 & \textbf{73.16} & \textbf{78.24} & \textbf{84.36} & \textbf{66.48} & \textbf{58.91} & \textbf{66.96} & 58.60 & \textbf{62.50} \\
            \hline \hline
            IN Pre-train & 20\%                             & 99.53 & 36.33 & 37.57 & 49.56 & 32.73 & 33.11 & 47.01 & 42.29 & 29.79 & 34.31 & 39.52 & 29.84 & 33.40 & 46.71 & 55.73 & 60.20 & 54.27 & 58.36 & 64.67 & 68.74 & 81.83 & 49.21 & 43.69 & 48.51 & 44.49 & 48.86 \\
            CMC$^\dagger$~\cite{tian2020contrastive} & 20\% & 99.58 & 44.54 & 49.05 & 54.20 & 42.95 & \textbf{47.77} & 52.52 & 50.89 & 40.65 & 40.08 & 49.67 & 39.46 & 39.85 & 56.26 & 68.09 & 70.69 & 66.45 & 69.76 & 69.23 & 73.95 & 81.87 & 62.20 & 52.63 & 61.35 & 51.45 & 57.41 \\
            MMV$^\dagger$~\cite{alayrac2020self} & 20\%     & 99.58 & 44.94 & 51.80 & 56.68 & 38.96 & 45.12 & 53.29 & 50.16 & 40.35 & 40.29 & 48.96 & 39.99 & 39.31 & 56.14 & 68.21 & 72.31 & 66.00 & 70.07 & 68.50 & 73.38 & 81.62 & 61.54 & 52.27 & 61.23 & 51.28 & 57.28 \\
            Ours$^\dagger$ & 20\%                           & \textbf{99.60} & \textbf{48.34} & \textbf{54.49} & \textbf{58.57} & \textbf{43.35} & 47.57 & \textbf{55.65} & \textbf{55.46} & \textbf{45.50} & \textbf{46.55} & \textbf{55.34} & \textbf{44.65} & \textbf{46.47} & \textbf{61.25} & \textbf{71.12} & \textbf{73.61} & \textbf{69.45} & \textbf{71.37} & \textbf{71.44} & \textbf{76.74} & \textbf{83.52} & \textbf{64.86} & \textbf{56.40} & \textbf{64.23} & \textbf{55.73} & \textbf{60.85} \\
            \hline \hline
            IN Pre-train & 10\%                             & 99.52 & 31.06 & 30.86 & 42.62 & 31.49 & 28.61 & 39.59 & 40.24 & 24.58 & 28.15 & 38.36 & 25.25 & 26.65 & 47.38 & 50.07 & 53.83 & 49.90 & 52.64 & 63.70 & 68.21 & 80.72 & 44.98 & 35.24 & 44.19 & 35.80 & 44.55 \\
            CMC$^\dagger$~\cite{tian2020contrastive} & 10\% & 99.55 & 38.50 & 45.93 & 48.87 & 42.27 & 42.37 & 47.59 & 48.30 & 37.07 & 36.52 & 47.46 & 36.72 & 36.07 & 55.22 & 65.35 & 66.68 & 65.23 & 64.92 & 67.31 & 72.76 & 81.25 & 59.06 & 48.16 & 58.33 & 47.14 & 54.35 \\
            MMV$^\dagger$~\cite{alayrac2020self} & 10\%     & 99.54 & 38.12 & 45.86 & 50.23 & 37.23 & 40.96 & 48.24 & 47.57 & 36.17 & 36.33 & 46.99 & 36.13 & 35.34 & 55.36 & 64.36 & 66.91 & 63.40 & 64.91 & 67.40 & 72.44 & 80.88 & 58.72 & 47.61 & 58.71 & 46.98 & 53.86 \\
            Ours$^\dagger$ & 10\%                           & \textbf{99.57} & \textbf{43.01} & \textbf{49.27} & \textbf{53.44} & \textbf{46.22} & \textbf{46.13} & \textbf{52.20} & \textbf{53.68} & \textbf{40.73} & \textbf{43.25} & \textbf{54.63} & \textbf{41.18} & \textbf{43.80} & \textbf{59.88} & \textbf{67.81} & \textbf{69.59} & \textbf{69.08} & \textbf{67.91} & \textbf{69.02} & \textbf{75.36} & \textbf{82.30} & \textbf{62.99} & \textbf{51.10} & \textbf{63.35} & \textbf{51.42} & \textbf{58.28} \\
            \hline \hline
            IN Pre-train & 1\%                             & 99.07 & 0.00 & 0.00 & 8.97 & 0.00 & 0.00 & 6.92 & 0.00 & 0.00 & 0.00 & 0.00 & 0.00 & 0.00 & 0.00 & 19.38 & 24.79 & 19.34 & 23.11 & 43.39 & 48.80 & 72.37 & 0.00 & 0.00 & 0.00 & 0.00 & 14.65 \\
            CMC$^\dagger$~\cite{tian2020contrastive} & 1\% & 99.08 & 0.00 & 0.00 & 17.72 & 0.00 & 0.00 & 14.43 & 0.00 & 0.00 & 0.00 & 0.00 & 0.00 & 0.00 & 33.14 & 26.55 & 24.95 & 25.94 & 22.73 & 48.98 & 52.21 & 72.88 & 2.65 & 0.00 & 2.99 & 0.00 & 17.77 \\
            MMV$^\dagger$~\cite{alayrac2020self} & 1\%     & 99.10 & 0.00 & 0.00 & 12.83 & 0.00 & 0.00 & 10.97 & 0.00 & 0.00 & 0.00 & 0.00 & 0.00 & 0.00 & 23.62 & 26.53 & 24.72 & 26.40 & 24.54 & 51.18 & 53.27 & 72.29 & 8.15 & 0.00 & 7.96 & 0.00 & 17.66 \\
            Ours$^\dagger$ & 1\%                           & \textbf{99.15} & 0.00 & 0.00 & \textbf{18.29} & 0.00 & 0.00 & \textbf{16.99} & \textbf{0.07} & 0.00 & 0.00 & \textbf{0.07} & 0.00 & 0.00 & \textbf{46.56} & \textbf{28.73} & \textbf{27.59} & \textbf{29.13} & \textbf{26.38} & \textbf{58.59} & \textbf{61.46} & \textbf{74.41} & \textbf{16.32} & 0.00 & \textbf{15.65} & 0.00 & \textbf{20.78} \\
            \hline \hline
            Ablation1 & 10\% & 99.55 & 38.50 & 45.93 & 48.87 & 42.27 & 42.37 & 47.59 & 48.30 & 37.07 & 36.52 & 47.46 & 36.72 & 36.07 & 55.22 & 65.35 & 66.68 & 65.23 & 64.92 & 67.31 & 72.76 & 81.25 & 59.06 & 48.16 & 58.33 & 47.14 & 54.35 \\
            Ablation2 & 10\% & 99.56 & 40.06 & 45.94 & 50.88 & 40.18 & 43.91 & 49.06 & 50.41 & 36.73 & 37.65 & 49.50 & 37.33 & 36.94 & 57.19 & 66.32 & 68.69 & 67.28 & 66.98 & 68.08 & 73.37 & 81.95 & 60.25 & 48.45 & 59.80 & 47.54 & 55.36 \\
            Ablation3 & 10\% & \textbf{99.57} & \textbf{44.49} & 46.38 & 52.15 & 44.00 & 43.85 & 50.25 & 51.73 & 37.19 & 38.12 & 51.30 & 37.24 & 37.61 & 57.38 & 67.16 & 69.08 & 66.91 & 67.32 & \textbf{69.42} & 74.45 & 82.12 & 61.36 & 49.91 & 60.69 & 48.97 & 56.35 \\
            Ablation4 & 10\% & \textbf{99.57} & 43.01 & \textbf{49.27} & \textbf{53.44} & \textbf{46.22} & \textbf{46.13} & \textbf{52.20} & \textbf{53.68} & \textbf{40.73} & \textbf{43.25} & \textbf{54.63} & \textbf{41.18} & \textbf{43.80} & \textbf{59.88} & \textbf{67.81} & \textbf{69.59} & \textbf{69.08} & \textbf{67.91} & 69.02 & \textbf{75.36} & \textbf{82.30} & \textbf{62.99} & \textbf{51.10} & \textbf{63.35} & \textbf{51.42} & \textbf{58.28} \\
            \Xhline{1pt}
        \end{tabular}
    }
    \end{center}
    \label{tab:human36mdetails}
\end{sidewaystable}

\clearpage
\begin{sidewaystable}[!htbp]
    \vspace{80pt}
    \caption{Detailed Human Parsing Results on NTURGBD-Parsing-4K. `Ratio' stands for the ratio of training data for downstream tasks transfer. \textsuperscript{*} randomly initializes the model before pre-training. $^\dagger$ initializes the model by ImageNet pre-train before pre-training. `Ablation1' is `Sample-level Mod-invariant'; `Abation2' is `+ Hard Dense Intra-sample'; `Ablation3' is `+ Soft Dense Intra-sample'; `Ablation4' is `+ Sparse Structure-aware'. All results in [\%].}
    \vspace{-10pt}
    \begin{center}
    \tiny{
        \addtolength{\tabcolsep}{-3.9pt}
        \begin{tabular}{l|c|ccccccccccccccccccccccccc|c}
            \Xhline{1pt}
            Methods & Ratio & bg & right hip & right knee & right foot & left hip & left knee & left foot & left shoulder & left elbow & left hand & right shoulder & right elbow & right hand & crotch & right thigh & right calf & left thigh & left calf & lower spine & upper spine & head & left arm & left forearm & right arm & right forearm & mIoU \\
            \hline \hline
            IN Pre-train & 100\%                             & 99.24 & 21.99 & 19.79 & 39.62 & 23.95 & 20.98 & 39.61 & 22.61 & 14.14 & 22.11 & 23.05 & 12.24 & 22.52 & 25.66 & 47.00 & 46.85 & 46.81 & 48.53 & 53.26 & 61.51 & 61.11 & 43.45 & 36.13 & 46.50 & 38.56 & 37.49 \\
            CMC$^\dagger$~\cite{tian2020contrastive} & 100\% & 99.26 & 22.50 & 19.49 & 40.15 & 24.81 & 20.65 & 39.96 & \textbf{24.64} & 14.68 & 21.79 & \textbf{25.50} & 13.04 & 23.46 & \textbf{26.12} & 48.93 & 46.21 & 49.28 & 47.68 & 54.26 & 62.52 & 59.24 & 45.02 & 37.22 & 48.32 & 40.25 & 38.20 \\
            MMV$^\dagger$~\cite{alayrac2020self} & 100\%     & 99.23 & 22.64 & 19.68 & 39.02 & 24.66 & 21.38 & 38.77 & 24.40 & 13.92 & 22.36 & 25.15 & 12.83 & 23.65 & 25.78 & 47.87 & 46.62 & 48.12 & 48.45 & 53.32 & 62.46 & 59.89 & 45.03 & 37.80 & 48.17 & 40.98 & 38.09 \\
            Ours$^\dagger$ & 100\%                           & \textbf{99.32} & \textbf{22.95} & \textbf{21.25} & \textbf{41.26} & \textbf{25.70} & \textbf{22.59} & \textbf{40.99} & 24.47 & \textbf{15.17} & \textbf{23.61} & 25.11 & \textbf{14.26} & \textbf{24.87} & 25.75 & \textbf{49.46} & \textbf{47.90} & \textbf{49.97} & \textbf{49.88} & \textbf{54.45} & \textbf{62.88} & \textbf{61.97} & \textbf{47.24} & \textbf{39.06} & \textbf{50.48} & \textbf{42.29} & \textbf{39.32} \\
            \hline \hline
            IN Pre-train & 20\%                             & 99.13 & 12.39 & 17.07 & 31.82 & 14.08 & 19.28 & 32.72 & 13.68 &  2.61 & 12.58 & 14.59 &  2.96 & 10.87 & 18.16 & 35.01 & 33.36 & 38.41 & 35.82 & 46.81 & 54.45 & 57.49 & 31.95 & 23.39 & 32.65 & 22.60 & 28.56 \\
            CMC$^\dagger$~\cite{tian2020contrastive} & 20\% & 98.98 & \textbf{13.16} & 14.06 & 29.82 & \textbf{16.28} & 16.13 & 30.99 & 17.65 & 10.09 & 14.69 & 17.99 &  8.23 & 15.92 & 18.08 & 38.79 & 34.57 & 41.26 & 36.41 & 46.21 & 54.68 & 54.60 & 37.80 & 26.75 & 39.11 & 27.66 & 30.40 \\
            MMV$^\dagger$~\cite{alayrac2020self} & 20\%     & 99.03 & 12.92 & 16.15 & 30.81 & 15.62 & 18.80 & 32.06 & 17.75 &  7.46 & 15.27 & 18.73 &  5.52 & 15.72 & \textbf{19.48} & 38.29 & 34.88 & 40.95 & 37.82 & 45.78 & 54.42 & 55.21 & 36.83 & 25.66 & 37.85 & 27.15 & 30.41 \\
            Ours$^\dagger$ & 20\%                           & \textbf{99.40} & 12.18 & \textbf{18.25} & \textbf{36.74} & 15.35 & \textbf{21.15} & \textbf{38.91} & \textbf{18.78} & \textbf{11.99} & \textbf{20.92} & \textbf{20.31} & \textbf{10.99} & \textbf{20.84} & 18.05 & \textbf{44.73} & \textbf{44.45} & \textbf{47.86} & \textbf{46.87} & \textbf{51.41} & \textbf{59.57} & \textbf{65.76} & \textbf{42.43} & \textbf{31.07} & \textbf{44.67} & \textbf{32.65} & \textbf{35.01} \\
            \hline \hline
            Ablation1 & 20\% & 99.13 & 12.39 & 17.07 & 31.82 & 14.08 & 19.28 & 32.72 & 13.68 & 2.61 & 12.58 & 14.59 & 2.96 & 10.87 & 18.16 & 35.01 & 33.36 & 38.41 & 35.82 & 46.81 & 54.45 & 57.49 & 31.95 & 23.39 & 32.65 & 22.60 & 28.56 \\
            Ablation2 & 20\% & 99.12 & 12.85 & 13.79 & 32.69 & \textbf{16.99} & 16.20 & 33.29 & 17.38 & 8.73 & 15.12 & 19.13 & 7.67 & 16.71 & \textbf{19.07} & 39.46 & 36.15 & 42.08 & 38.08 & 48.69 & 55.53 & 58.94 & 38.49 & 27.62 & 40.04 & 27.55 & 31.26 \\
            Ablation3 & 20\% & 99.22 & \textbf{13.40} & 14.18 & 31.37 & 16.46 & 17.71 & 32.88 & \textbf{19.72} & 9.33 & 16.13 & 20.19 & 8.95 & 17.49 & 17.48 & 39.12 & 37.41 & 41.55 & 39.13 & 49.67 & 58.23 & 61.41 & 39.71 & 30.95 & 41.16 & 32.13 & 32.20 \\
            Ablation4 & 20\% & \textbf{99.40} & 12.18 & \textbf{18.25} & \textbf{36.74} & 15.35 & \textbf{21.15} & \textbf{38.91} & 18.78 & \textbf{11.99} & \textbf{20.92} & \textbf{20.31} & \textbf{10.99} & \textbf{20.84} & 18.05 & \textbf{44.73} & \textbf{44.45} & \textbf{47.86} & \textbf{46.87} & \textbf{51.41} & \textbf{59.57} & \textbf{65.76} & \textbf{42.43} & \textbf{31.07} & \textbf{44.67} & \textbf{32.65} & \textbf{35.01} \\
            \hline \hline
            From Scratch w/ PN++ & 20\%                                      & - & 21.43 & 30.84 & 67.53 & 21.52 & 29.85 & 66.76 & 32.82 & 23.17 & 38.37 & 36.74 & 23.42 & 36.23 & 25.19 & 55.25 & 60.25 & 55.11 & 60.37 & 53.34 & 65.85 & 88.22 & 50.77 & 47.05 & 52.70 & 45.88 & 45.36 \\
            CMC\textsuperscript{*}~\cite{tian2020contrastive} w/ PN++ & 20\% & - & \textbf{24.12} & 32.89 & 73.11 & 23.84 & \textbf{32.67} & \textbf{73.20} & 33.43 & 27.55 & 44.62 & \textbf{38.40} & 27.59 & 42.11 & 26.55 & 57.82 & 65.60 & 57.73 & 65.02 & 54.53 & 66.31 & 89.16 & 55.04 & 51.00 & 56.66 & 50.82 & 48.74 \\
            Ours\textsuperscript{*} w/ PN++ & 20\%                           & - & 23.96 & \textbf{32.90} & \textbf{73.30} & \textbf{24.16} & 32.44 & 73.10 & \textbf{34.81} & \textbf{29.54} & \textbf{45.43} & 37.79 & \textbf{28.16} & \textbf{42.89} & \textbf{27.83} & \textbf{58.25} & \textbf{66.16} & \textbf{58.64} & \textbf{65.51} & \textbf{55.60} & \textbf{66.92} & \textbf{89.51} & \textbf{56.33} & \textbf{53.05} & \textbf{57.87} & \textbf{52.29} & \textbf{49.43} \\
            \Xhline{1pt}
        \end{tabular}
    }
    \end{center}
    \label{tab:nturgbddetails}
\end{sidewaystable}

\begin{sidewaystable}[!htbp]
    \vspace{-20pt}
    \caption{Detailed Cross-Modality Supervision and Missing-Modality Inference Results on NTURGBD-Parsing-4K. All results in [\%].}
    \vspace{-10pt}
    \begin{center}
    \tiny{
        \addtolength{\tabcolsep}{-4pt}
        \begin{tabular}{l|c|ccccccccccccccccccccccccc|c}
            \Xhline{1pt}
            Methods & Setting & bg & right hip & right knee & right foot & left hip & left knee & left foot & left shoulder & left elbow & left hand & right shoulder & right elbow & right hand & crotch & right thigh & right calf & left thigh & left calf & lower spine & upper spine & head & left arm & left forearm & right arm & right forearm & mIoU \\
            \hline \hline
            No Contrastive & RGB$\rightarrow$Depth                 & 92.87 &  0.00 &  0.00 &  0.00 &  0.00 &  0.00 &  0.00 &  0.00 &  0.00 &  0.22 &  0.00 &  0.00 &  1.08 &  0.00 &  0.00 &  0.25 &  0.00 &  0.00 &  0.00 &  0.00 &  2.62 &  0.00 &  0.45 &  0.00 &  1.11 & 3.94 \\
            CMC~\cite{tian2020contrastive} & RGB$\rightarrow$Depth & 89.79 &  0.00 &  0.00 &  0.00 &  0.00 &  0.02 &  0.01 &  0.00 &  1.51 &  0.79 &  0.00 &  0.11 &  0.43 &  0.00 &  0.00 &  0.00 &  0.01 &  0.00 &  1.02 &  0.20 &  0.42 &  1.64 &  0.37 &  0.03 &  0.14 & 3.86 \\
            Ours & RGB$\rightarrow$Depth                           & \textbf{96.78} & \textbf{12.50} & \textbf{23.26} & \textbf{37.60} & \textbf{16.45} & \textbf{21.43} & \textbf{40.51} & \textbf{22.41} & \textbf{19.59} & \textbf{22.86} & \textbf{21.35} & \textbf{17.19} & \textbf{25.67} & \textbf{17.40} & \textbf{36.48} & \textbf{35.62} & \textbf{37.43} & \textbf{34.04} & \textbf{49.83} & \textbf{59.37} & \textbf{61.07} & \textbf{33.91} & \textbf{24.55} & \textbf{36.15} & \textbf{26.37} & \textbf{33.19} \\
            \hline \hline
            No Contrastive & Depth$\rightarrow$RGB                 & 91.79 &  0.00 &  0.00 &  0.00 &  0.00 &  0.00 &  0.00 &  0.00 &  0.00 &  0.23 &  0.00 &  0.00 &  0.07 &  0.00 &  0.01 &  0.00 &  0.00 &  0.00 &  0.10 &  0.00 &  0.02 &  0.00 &  0.50 &  0.00 &  0.02 & 3.71 \\
            CMC~\cite{tian2020contrastive} & Depth$\rightarrow$RGB & 91.96 &  0.00 &  0.00 &  0.00 &  0.00 &  0.00 &  0.01 &  0.00 &  0.68 &  0.32 &  0.00 &  0.00 &  0.21 &  0.00 &  0.00 &  0.00 &  0.00 &  0.00 &  0.46 &  0.23 &  0.05 &  0.49 &  1.84 &  0.00 &  0.01 & 3.85 \\
            Ours & Depth$\rightarrow$RGB                           & \textbf{95.15} & \textbf{13.70} & \textbf{16.04} & \textbf{28.70} & \textbf{16.59} & \textbf{12.15} & \textbf{28.12} & \textbf{18.11} & \textbf{12.78} & \textbf{10.51} & \textbf{21.06} & \textbf{11.12} & \textbf{14.46} & \textbf{11.02} & \textbf{33.71} & \textbf{30.90} & \textbf{34.03} & \textbf{22.93} & \textbf{42.62} & \textbf{50.99} & \textbf{56.12} & \textbf{23.51} & \textbf{19.89} & \textbf{27.29} & \textbf{18.50} & \textbf{26.80} \\
            \hline \hline
            No Contrastive & Only RGB                              & 93.55 &  8.97 &  6.66 &  0.42 &  4.28 &  0.75 &  0.02 &  0.98 &  1.19 & 15.22 &  0.28 &  2.69 & 23.56 &  0.08 &  0.30 & 12.35 &  0.07 &  0.88 & 25.48 &  5.27 & 57.54 & 12.63 & 26.38 &  7.73 & 28.90 & 13.45 \\
            CMC~\cite{tian2020contrastive} & Only RGB              & 93.80 &  0.00 & 11.94 & 47.69 &  0.00 & 12.00 & 38.76 & 21.43 &  0.01 &  9.58 & 24.32 & 13.56 & 15.38 &  1.15 &  1.84 & 32.30 &  1.07 & 18.12 &  0.59 &  3.13 & 36.86 & 27.04 & 15.63 & 42.37 & 21.90 & 19.62 \\
            Ours & Only RGB                                        & \textbf{97.80} & \textbf{29.12} & \textbf{27.49} & \textbf{57.93} & \textbf{27.09} & \textbf{26.41} & \textbf{57.31} & \textbf{28.22} & \textbf{27.62} & \textbf{32.70} & \textbf{29.51} & \textbf{26.16} & \textbf{33.41} & \textbf{25.66} & \textbf{52.42} & \textbf{49.60} & \textbf{52.38} & \textbf{54.91} & \textbf{52.40} & \textbf{52.75} & \textbf{75.69} & \textbf{47.98} & \textbf{41.69} & \textbf{50.68} & \textbf{40.08} & \textbf{43.88} \\
            \hline \hline
            No Contrastive & Only Depth                 & 96.46 & \textbf{27.81} &  7.16 &  1.46 & \textbf{33.36} & 10.60 &  2.30 & 28.64 &  4.72 &  1.27 & 11.49 &  0.11 &  7.86 & 26.16 & 33.67 & 39.55 & 33.21 & 27.40 & 46.05 & 63.16 & 47.50 & 25.38 &  9.99 & 19.89 &  5.04 & 24.41 \\
            CMC~\cite{tian2020contrastive} & Only Depth & 94.81 &  7.25 &  1.08 &  0.06 &  6.82 &  0.05 &  0.10 & 25.47 &  3.11 &  2.95 & 20.80 &  0.39 &  2.73 & 26.79 & 17.11 & 17.84 & 33.96 &  4.58 & 10.64 & 11.32 & 37.01 & 41.41 & 11.08 & 33.69 &  3.52 & 16.58 \\
            Ours & Only Depth                           & \textbf{97.89} & 23.09 & \textbf{32.97} & \textbf{54.55} & 26.44 & \textbf{32.55} & \textbf{57.28} & \textbf{34.08} & \textbf{19.83} & \textbf{25.23} & \textbf{32.33} & \textbf{21.33} & \textbf{28.71} & \textbf{30.57} & \textbf{54.91} & \textbf{59.33} & \textbf{53.16} & \textbf{59.89} & \textbf{56.50} & \textbf{65.09} & \textbf{61.87} & \textbf{49.43} & \textbf{36.47} & \textbf{50.34} & \textbf{35.75} & \textbf{43.98} \\
            \Xhline{1pt}
        \end{tabular}
    }
    \end{center}
    \label{tab:appdetails}
\end{sidewaystable}

\clearpage
\clearpage
\clearpage

\end{document}